\documentclass[11pt]{article}

\usepackage[final]{acl}

\usepackage{times}
\usepackage{latexsym}

\usepackage[T1]{fontenc}

\usepackage[utf8]{inputenc}

\usepackage{microtype}

\usepackage{inconsolata}

\usepackage{graphicx}

\usepackage{url}
\usepackage{booktabs}
\usepackage{amsfonts}
\usepackage{nicefrac}
\usepackage{microtype}
\usepackage{graphicx}
\usepackage{algorithm}
\usepackage{algorithmic}
\usepackage{amsmath,amssymb,bm}
\usepackage{xcolor}
\usepackage{wrapfig}
\usepackage{caption}
\usepackage{enumitem}
\usepackage{tabularx}
\usepackage[most]{tcolorbox}
\tcbuselibrary{listings}
\usepackage{natbib}
\usepackage{colortbl}
\usepackage{array}
\usepackage{multirow}
\usepackage[normalem]{ulem}
\usepackage{titlesec}
\usepackage{xcolor}

\usepackage{etoc}
\usepackage{tocloft}

\usepackage{subcaption}
\usepackage{enumitem}

\definecolor{first}{HTML}{A8D4FF}   
\definecolor{second}{HTML}{C8E4FF}  
\definecolor{third}{HTML}{E8F4FF}   
\definecolor{better}{HTML}{008800}  
\definecolor{worse}{HTML}{CC0000}   

\usepackage{xspace}

\definecolor{yellowgreen}{rgb}{0.4,0.8,0.1}

\newtcolorbox{promptbox}{
  colback=gray!5,
  colframe=black!50,
  boxrule=0.5pt,
  arc=2pt,
  left=6pt, right=6pt, top=6pt, bottom=6pt,
  width=0.98\linewidth,
  breakable,
  pad at break=6pt,
}

\newcommand{\customfootnotetext}[1]{%
  \begingroup
    \renewcommand{\thefootnote}{}
    \footnotetext{#1}%
  \endgroup
}

\title{\raisebox{-0.3\height}{\includegraphics[width=1.2cm]{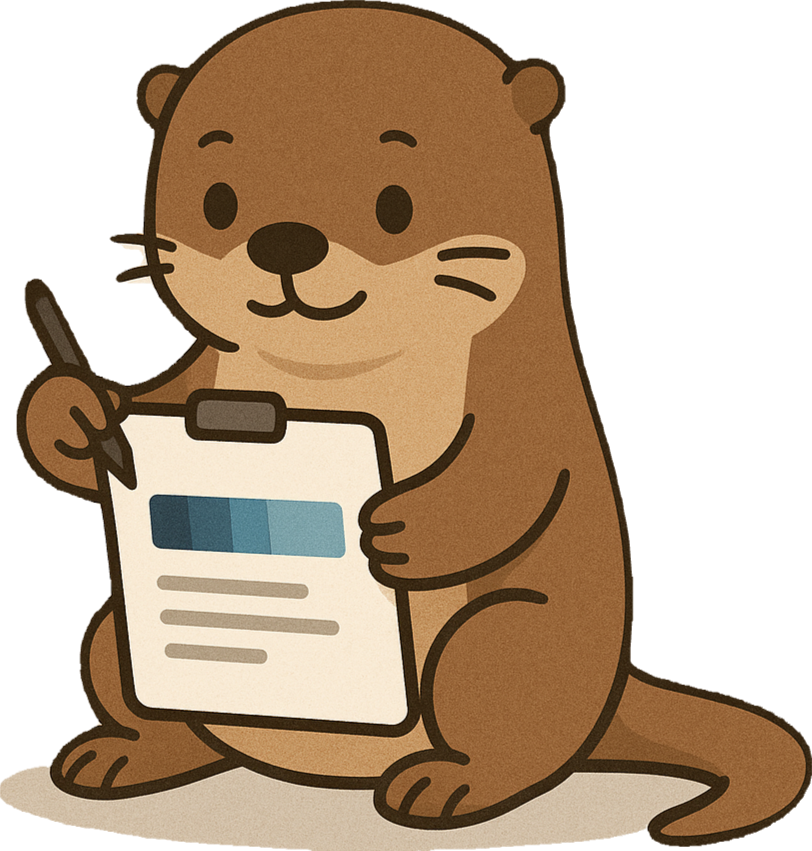}}~ProgressLM: Towards Progress Reasoning in\\ Vision-Language Models}

\usepackage{fontawesome}
\definecolor{projectlink}{HTML}{0366D6}

\author{
{\mdseries Jianshu Zhang}$^{1*}$ \quad {\mdseries Chengxuan Qian}$^{2*}$ \quad {\mdseries Haosen Sun}$^{1}$\\[0.1em]
{\mdseries Haoran Lu}$^{1}$ \quad {\mdseries Dingcheng Wang}$^{1}$ \quad {\mdseries Letian Xue}$^{1}$ \quad {\mdseries Han Liu}$^{1\dagger}$\\[0.3em]
$^{1}$Northwestern University \quad
$^{2}$University of California, Santa Barbara\\[0.2em]
{\small \texttt{sterzhang@u.northwestern.edu}, \texttt{chengxuanqian@ucsb.edu}, \texttt{hanliu@northwestern.edu}}\\[0.5em]
{\color{projectlink}
\faGlobe\ \href{https://progresslm.github.io/ProgressLM/}{Website}
\quad
\faGithub\ \href{https://github.com/ProgressLM/ProgressLM}{Code}
\quad
\faCube\ \href{https://huggingface.co/collections/Raymond-Qiancx/progresslm}{Model}
\quad
\faDatabase\ \href{https://huggingface.co/datasets/Raymond-Qiancx/ProgressLM-Dataset}{Dataset}
}
}

\begin{document}
\maketitle
\customfootnotetext{$^{*}$Equal contribution. $^{\dagger}$Corresponding author.}
\begin{abstract}
Estimating task progress requires long-horizon and dynamic reasoning, going beyond static visual perception.
Although Vision-Language Models (VLMs) excel at describing what is visible in a single observation, it remains unclear whether they can infer \emph{how far a task has progressed} from partial information.
To study this question, we introduce \textbf{\textsc{Progress-Bench}}, a benchmark with over 3K instances for evaluating progress reasoning from a single observation.
We further examine a human-inspired two-stage paradigm that combines episodic retrieval with mental simulation.
We instantiate this paradigm through both training-free prompting and a training-based approach using the automatically curated \textsc{ProgressLM-45K} dataset.
Experiments on 14 VLMs show that most models struggle with reliable progress estimation, and that training-free reasoning provides only limited and model-dependent benefits.
In contrast, the training-based \textsc{ProgressLM-3B} achieves consistent improvements in accuracy, robustness to viewpoint variation, and handling of unanswerable cases, despite its small scale.
Additional analyses reveal common failure patterns in existing VLMs and clarify when and why progress reasoning succeeds or fails.
\end{abstract}

\section{Introduction}

Given an observation from an intermediate moment of a task, most Vision Language Models (VLMs)~\citep{hurst2024gpt, bai2025qwen2, wang2025internvl3} can accurately describe what is visible.
However, estimating \emph{how much of the task has been completed} is fundamentally different, as it requires long-horizon, dynamic reasoning beyond snapshot-level perception.

Prior work on progress estimation either relies on task-specific regression models~\citep{yang2024rank2reward, chen2025sarmstageawarerewardmodeling}, or infers progress indirectly via surrogate objectives such as trajectory reordering~\citep{ma2024vision} or pairwise comparison~\citep{zhai2025vision}.
This raises a fundamental question: \emph{can VLMs acquire progress estimation as a general reasoning capability from a single observation?}
To systematically study this problem, we introduce \textbf{\textsc{Progress-Bench}}, a benchmark built on the robotic manipulation domain~\citep{wu2025robomind}, where task execution follows clear and temporally ordered progressions.
Each instance consists of a task demonstration and a single observation, and the model is required to predict a numerical progress score.
The benchmark is designed to probe progress reasoning along three main axes: demonstration modality (vision vs.\ text), viewpoint correspondence (same-view vs.\ cross-view between the demonstration and the observation), and answerability.

\begin{figure*}[h]
    \centering
    \includegraphics[width=1\linewidth]{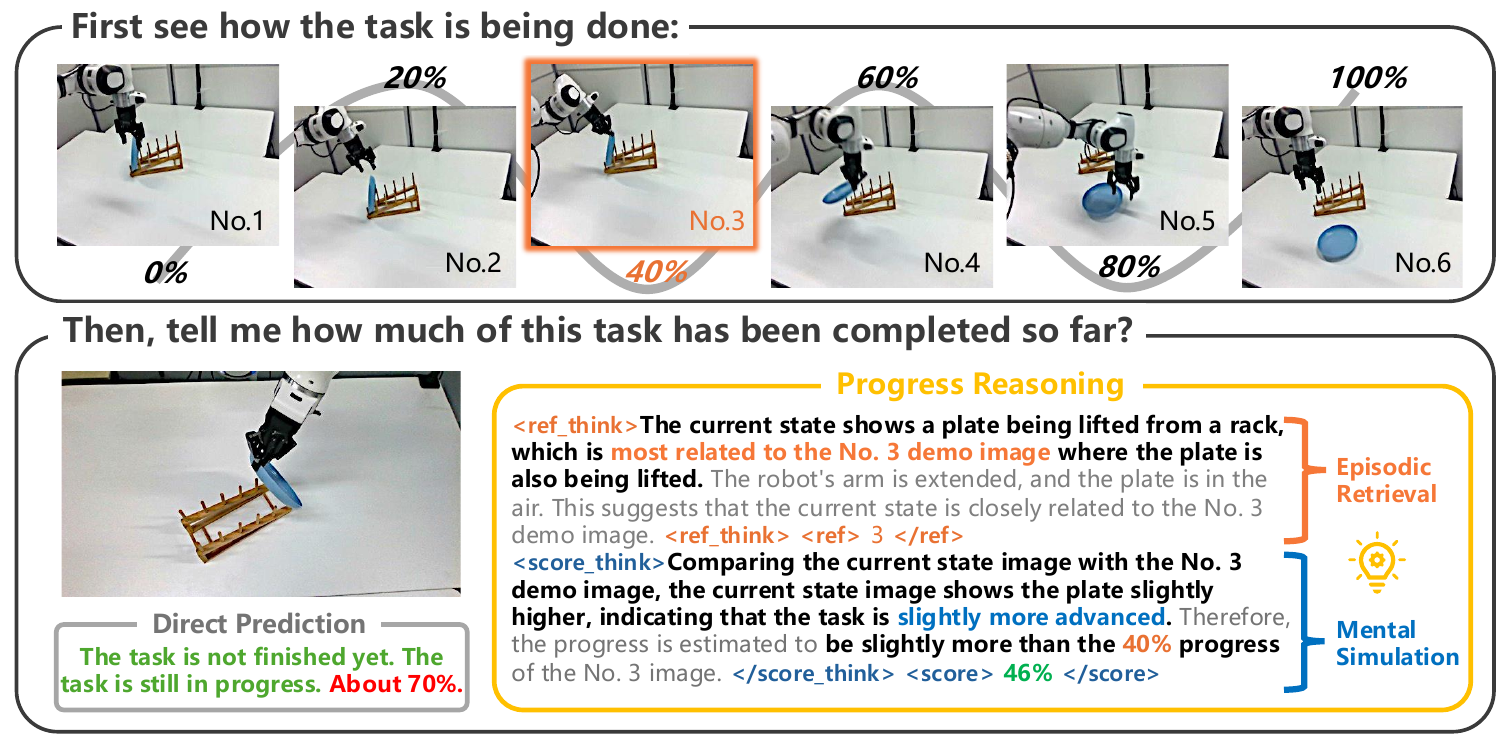}
    \caption{
Given a task demonstration and a single observation, direct prediction can often judge whether the task is unfinished, but struggles to assign a well-calibrated progress score.
Progress reasoning instead follows a coarse-to-fine process:
it first performs \emph{episodic retrieval} to coarsely locate the observation along the demonstrated task,
then applies \emph{mental simulation} to imagine the transition from the retrieved anchor to the current observation, yielding accurate and interpretable progress estimation.
}
    \vspace{-1em}
    \label{fig:teaser}
\end{figure*}

Beyond benchmarking existing models, we further ask: \textbf{How, then, can progress reasoning be effectively learned?}
Humans excel at progress estimation by interpreting task execution as a continuous process that combines \textit{episodic retrieval} to locate a coarse anchor along the task trajectory, and \textit{mental simulation} to reason about how the task state evolves from this anchor toward the current observation~\citep{schacter2008episodic}.
Inspired by this process, we first explore \textbf{training-free} prompting strategies that explicitly encourage VLMs to follow this two-stage reasoning pattern as shown in Figure~\ref{fig:teaser}.
To further endow models with robust progress reasoning ability, we explore a \textbf{training-based} approach and automatically construct a dataset named \textsc{\textbf{ProgressLM-45K}}, with 25K chain-of-thought samples for supervised cold-start and 20K used for reinforcement learning refinement, yielding a progress-reasoning-enhanced model, \textsc{\textbf{ProgressLM-3B}}.

Our experiments across 14 models show that VLMs struggle to estimate task progress reliably from a single observation. Direct prediction leads to strong sensitivity to demonstration modality and viewpoint changes, as well as poor handling of unanswerable cases. Training-free progress reasoning provides only conditional benefits. In contrast, training-based \textsc{ProgressLM-3B} yields consistent improvements even at small model scale. Further analysis reveals that different models' error patterns. We additionally examine when progress reasoning can be effective and why demonstration modality plays a critical role. 

Our main contributions are as follows:
\begin{itemize}[leftmargin=1.5em]
    \item We introduce \textsc{Progress-Bench}, a benchmark with over 3K instances for systematically explore whether VLMs can perform progress reasoning from a single observation under variations of demonstration modality, viewpoint, and answerability.

    \item We benchmark 14 VLMs on \textsc{Progress-Bench} and show that existing models exhibit limited and unstable progress reasoning, with strong sensitivity to modality and viewpoint changes, poor handling of unanswerable cases, and frequent collapse to coarse or heuristic predictions.

    \item We investigate how progress reasoning can be improved through a human-inspired two-stage paradigm.
    While training-free prompting yields only conditional gains, explicit training leads to \textsc{ProgressLM-3B}, which achieves performance comparable to or surpassing GPT-5 on \textsc{Progress-Bench}.
    Further analyses reveal common failure patterns in existing VLMs and clarify when and why progress reasoning succeeds or fails.
\end{itemize}

\begin{figure*}[t]
    \centering
    \includegraphics[width=\linewidth]{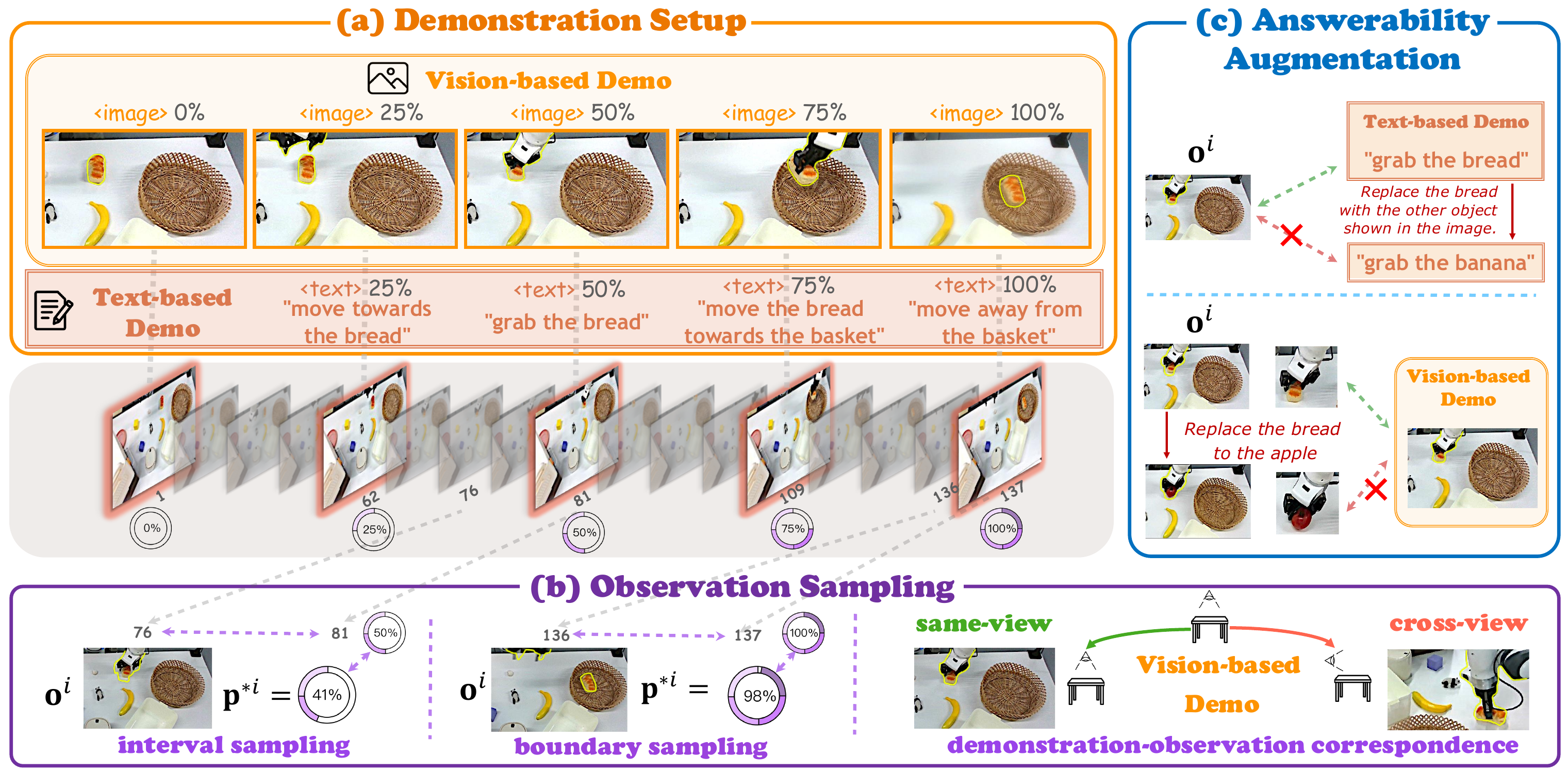}
    \caption{
Overview of \textsc{Progress-Bench}.
(a) \textbf{Demonstration setup} provides vision-based key-frame demonstrations or text-based step descriptions with progress annotations.
(b) \textbf{Observation sampling} selects observations between or near demonstration steps, with progress assigned by interpolation; vision-based settings include same-view and cross-view cases.
(c) \textbf{Answerability augmentation} creates unanswerable samples by introducing mismatches between demonstrations and observations.
}
    \label{fig:bench_pipeline}
\end{figure*}

\begin{figure*}[t]
    \centering
    \includegraphics[width=1\linewidth]{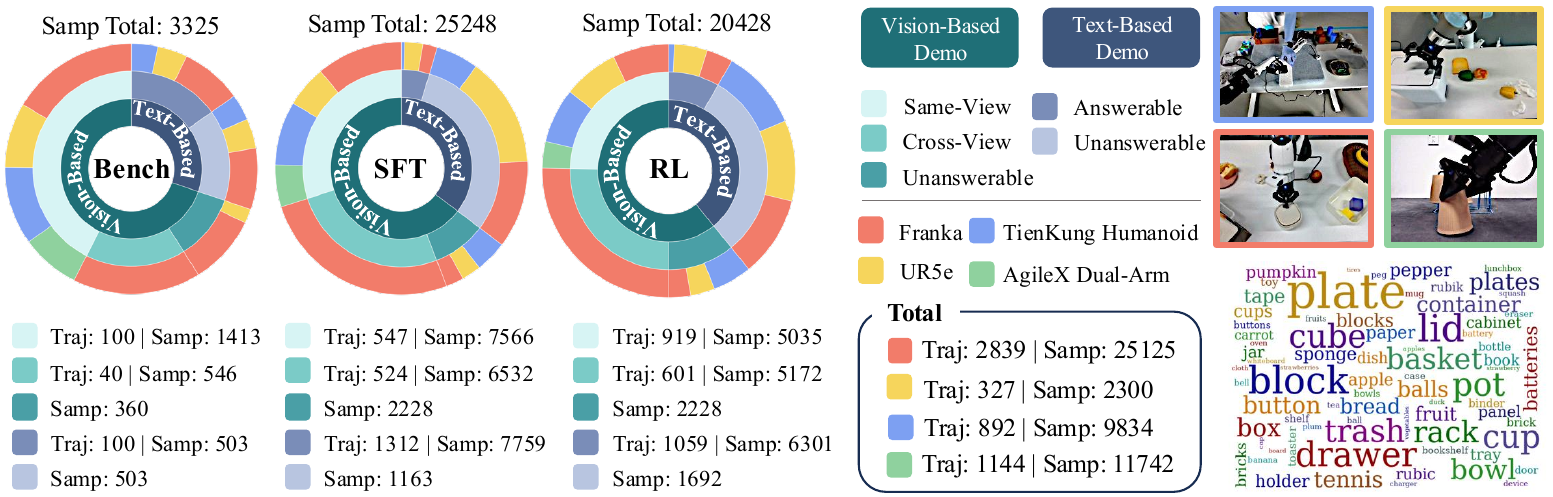}
    \caption{
Data statistics of \textsc{Progress-Bench} and \textsc{ProgressLM-45K} (25K SFT, 20K RL).
Traj and Samp indicate trajectory and sample counts; right panels show robot embodiments and object diversity.
}
\vspace{-1em}
    \label{fig:data_statistics}
\end{figure*}

\section{\textsc{Progress-Bench}}

\paragraph{Overview.}
\textsc{Progress-Bench} evaluates whether a model can infer task progress from a single observation by situating it within the temporal structure of an ongoing task, going beyond static perception toward progression reasoning.
Each instance consists of a task demonstration $D$ and an observation $o$ sampled from an intermediate execution stage.
The demonstration covers the full task from start to completion and is presented either as a sequence of images or as stepwise textual actions.
Given $(D,o)$, the model predicts a normalized progress score $p\in[0,100\%]$ indicating how far the task has advanced.
When the demonstration is insufficient or inconsistent with the observation, the model should instead output \texttt{N/A}.
To systematically probe this capability, \textsc{Progress-Bench} is constructed by varying demonstration modality, observation--demonstration viewpoint correspondence, and answerability, enabling controlled analysis of perception, temporal reasoning, and uncertainty handling.

\subsection{Benchmark Construction}

\paragraph{Demonstration Setup.}
We consider two demonstration modality types.
\textit{Vision-based} demonstrations consist of key frames from expert executions,
$\mathcal{D}_{v}=\{(\mathbf{f}_{a_j},\mathbf{p}_j)\}_{j=1}^{N}$, where each frame depicts a complete task state.
\textit{Text-based} demonstrations provide stepwise action descriptions,
$\mathcal{D}_{t}=\{(\mathbf{t}_j,\mathbf{p}_j)\}_{j=1}^{N}$.
Unlike vision-based demonstrations, text-based ones require integrating action semantics and accumulating implicit state changes, as intermediate states are not directly observable.

\paragraph{Observation Sampling.}
Given a demonstration $\mathcal{D}$ and its execution video $\mathbf{V}=\{\mathbf{f}_k\}_{k=1}^{T}$, we construct observation--progress pairs
$\mathcal{O}=\{(\mathbf{o}^i,\mathbf{p}^{*i})\}_{i=1}^{M}$.
The video is partitioned into $N{-}1$ segments between consecutive key steps.
For each segment, we sample intermediate frames at relative positions $\delta\in(0,1)$ and assign progress via linear interpolation,
$\mathbf{p}^{*}=\mathbf{p}_j+\delta(\mathbf{p}_{j+1}-\mathbf{p}_j)$.
We adopt two strategies: \textit{interval sampling} for uniform coverage of intermediate progress and \textit{boundary sampling} for finer resolution near step transitions.
To evaluate viewpoint robustness, we further distinguish \textit{same-view} and \textit{cross-view} settings in the vision-based modality, where observations are captured from the same or different camera viewpoints as the demonstration.

\paragraph{Answerability Augmentation.}
To assess uncertainty awareness, we explicitly introduce \textit{answerability}.
\textit{Answerable} samples correspond to coherent executions where progress can be inferred, while \textit{unanswerable} samples arise when the observation is inconsistent with the demonstration and the correct output is \texttt{N/A}.
Such cases are constructed by either modifying the demonstration or editing the observation while keeping the other fixed.

\paragraph{Data Details.}
\textit{Source}: We build \textsc{Progress-Bench} on RoboMind~\citep{wu2025robomind}, which provides standardized human teleoperation data with consistent sampling and temporally coherent trajectories. 
\textit{Quality Control}: We adopt a two-level labeling scheme, where discrete step-level progress defines the task structure and linear interpolation is applied only within each step for fine-grained supervision. 
A manual inspection of $100$ randomly sampled trajectories shows that $86\%$ are fully smooth, with the rest containing only minor frame-level discontinuities.
\textit{Statistics}: As shown in Figure~\ref{fig:data_statistics}, the benchmark comprises  $240$ task trajectories and $3325$ sampled observations.



\begin{algorithm}[t]
\caption{Two-Stage Progress Reasoning}
\label{alg:two_stage_progress}
\small
\begin{algorithmic}[1]
\STATE \textbf{Input:} Task Demonstration $\mathcal{D}$, current observation $\mathbf{o}$

\STATE \textbf{Stage 1: Episodic Retrieval}
\STATE Retrieve a step $j^\star$ in $\mathcal{D}$ that is most close to $\mathbf{o}$
\STATE Output \texttt{<ref\_think>} and \texttt{<ref>}

\STATE \textbf{Stage 2: Mental Simulation}
\STATE Use $j^\star$ as the anchor and only compare it against $\mathbf{o}$, inferring whether $\mathbf{o}$ is before, near, or beyond $j^\star$
\STATE Output \texttt{<score\_think>} and \texttt{<score>}
\end{algorithmic}
\end{algorithm}

\section{Towards Progress Reasoning}
We frame progress reasoning as a human-inspired two-stage process (Algorithm~\ref{alg:two_stage_progress}).
Given a demonstration and a partial observation, humans first perform \textit{episodic retrieval} to identify a representative reference step as a coarse anchor, and then apply \textit{mental simulation} to reason how the task state evolves from this anchor to the current observation.
This formulation treats progress estimation as reasoning over a latent task trajectory, rather than matching observations to fixed timestamps.

\subsection{Training-Free Approach}
We instantiate this two-stage reasoning via structured prompting without parameter updates.
The prompt enforces an explicit schema with four fields:
\texttt{<ref\_think>} (episodic retrieval reasoning),
\texttt{<ref>} (retrieved reference step),
\texttt{<score\_think>} (mental simulation),
and \texttt{<score>} (final progress estimate),
which the model follows at inference time.

\subsection{Training-Based Approach}
We further adopt a training-based approach to explicitly teach episodic retrieval and mental simulation.
We construct \textsc{ProgressLM-45K} from non-overlapping manipulation tasks with \textsc{Progress-Bench}, ensuring generalizable reasoning. Statistics are in Figure~\ref{fig:data_statistics}.

\paragraph{Cold-Start Supervised Fine-Tuning.}
We first perform supervised fine-tuning on \textsc{ProgressLM-25K-CoT} to internalize the two-stage reasoning pattern.
Each instance includes a demonstration $\mathcal{D}^i$, an observation $\mathbf{o}^i$, and a reasoning sequence $\mathbf{r}^{i*}$ containing ground-truth \texttt{<ref>} and \texttt{<score>}.
Using guided reasoning completion, the model generates the intermediate fields \texttt{<ref\_think>} and \texttt{<score\_think>}.
Training minimizes the autoregressive negative log-likelihood:
\begin{equation}
\mathcal{L}_{\text{SFT}}
= -\frac{1}{N}\sum_{i=1}^{N}
\log P_\theta(\mathbf{r}^{i*}\mid \mathcal{D}^i,\mathbf{o}^i).
\end{equation}

\paragraph{Reinforcement Learning.}
To improve robustness and calibration, we apply a second-stage reinforcement learning procedure based on GRPO~\citep{shao2024deepseekmath}:
\begin{equation}
\mathcal{L}_{\text{RL}}
= -\mathbb{E}_{\mathbf{r}\sim P_\theta(\mathbf{r}\mid \mathcal{D},\mathbf{o})}
\big[ R(\mathbf{r}) \big],
\end{equation}
where $R(\mathbf{r})=\alpha R_{\text{format}}+\beta R_{\text{ref}}+\gamma R_{\text{score}}$
encourages structured reasoning, accurate reference retrieval, and precise progress estimation.
Specifically, $R_{\text{ref}}$ and $R_{\text{score}}$ are defined as the negative relative errors between the predicted and ground-truth reference index and progress score, respectively.
We train on 20K samples with $\alpha\!:\!\beta\!:\!\gamma=1\!:\!6\!:\!3$.

\begin{table*}[t]
\centering
\caption{
Results on \textit{answerable} samples are reported using Normalized Score Error (NSE)$\downarrow$, Progress Rank Correlation (PRC)$\uparrow$, and Answerable False Rejection Rate (AFRR)$\downarrow$.
\colorbox{first}{Best}, \colorbox{second}{Second}, and \colorbox{third}{Third} results are highlighted; colored deltas indicate the effect of training-free reasoning (\textcolor{better}{better} or \textcolor{worse}{worse}).
}
\label{tab:main_result}
\setlength{\tabcolsep}{3pt}
\resizebox{\textwidth}{!}{
\begin{tabular}{lccccccccccccc}
\toprule
\multirow{2}{*}{\textbf{Model}} &
\multicolumn{3}{c}{\textbf{Vision-based Demo}} &
\multicolumn{3}{c}{\textbf{Text-based Demo}} &
\multicolumn{3}{c}{\textbf{Micro Avg}} &
\multicolumn{3}{c}{\textbf{Macro Avg}} \\
\cmidrule(lr){2-4} \cmidrule(lr){5-7} \cmidrule(lr){8-10} \cmidrule(lr){11-13}
& \textbf{NSE}$\downarrow$ & \textbf{PRC}$\uparrow$ & \textbf{AFRR}$\downarrow$
& \textbf{NSE}$\downarrow$ & \textbf{PRC}$\uparrow$ & \textbf{AFRR}$\downarrow$
& \textbf{NSE}$\downarrow$ & \textbf{PRC}$\uparrow$ & \textbf{AFRR}$\downarrow$
& \textbf{NSE}$\downarrow$ & \textbf{PRC}$\uparrow$ & \textbf{AFRR}$\downarrow$ \\
\midrule
GPT-5            & \cellcolor{second}18.9 {\scriptsize\textcolor{better}{-0.6}} & \cellcolor{second}89.4 {\scriptsize\textcolor{worse}{-0.2}} & 1.3 {\scriptsize\textcolor{better}{-0.2}}  & \cellcolor{second}23.6 {\scriptsize\textcolor{better}{-2.8}} & \cellcolor{second}55.8 {\scriptsize\textcolor{better}{+4.3}} & 7.0 {\scriptsize\textcolor{worse}{+0.4}}  & \cellcolor{second}19.9 {\scriptsize\textcolor{better}{-1.0}} & \cellcolor{second}82.5 {\scriptsize\textcolor{better}{+0.8}} & 2.5 {\scriptsize\textcolor{better}{-0.1}}  & \cellcolor{second}21.3 {\scriptsize\textcolor{better}{-1.7}} & \cellcolor{second}72.6 {\scriptsize\textcolor{better}{+2.0}} & 4.2 {\scriptsize\textcolor{worse}{+0.1}}  \\
GPT-5-mini       & 20.7 {\scriptsize\textcolor{worse}{+0.6}} & \cellcolor{third}87.7 {\scriptsize\textcolor{worse}{-0.6}} & \cellcolor{third}0.4 {\scriptsize\textcolor{better}{-0.3}}  & \cellcolor{first}21.1 {\scriptsize\textcolor{worse}{+1.3}} & \cellcolor{third}55.2 {\scriptsize\textcolor{better}{+1.8}} & 9.7 {\scriptsize\textcolor{better}{-5.1}}  & 20.8 {\scriptsize\textcolor{worse}{+0.7}} & \cellcolor{third}81.1 {\scriptsize\textcolor{worse}{-0.1}} & 2.3 {\scriptsize\textcolor{better}{-1.3}}  & \cellcolor{third}20.9 {\scriptsize\textcolor{worse}{+1.0}} & \cellcolor{third}71.4 {\scriptsize\textcolor{better}{+0.6}} & 5.1 {\scriptsize\textcolor{better}{-2.7}}  \\
Qwen2.5-VL-72B    & 27.0 {\scriptsize\textcolor{better}{-2.5}} & 60.0 {\scriptsize\textcolor{better}{+18.2}} & 5.9 {\scriptsize\textcolor{better}{-1.1}} & 38.9 {\scriptsize\textcolor{better}{-12.5}} & 41.6 {\scriptsize\textcolor{better}{+11.7}} & \cellcolor{first}0.0 {\scriptsize\textcolor{worse}{+29.4}}  & 29.4 {\scriptsize\textcolor{better}{-4.5}} & 56.2 {\scriptsize\textcolor{better}{+16.7}} & 4.7 {\scriptsize\textcolor{worse}{+4.9}} & 32.9 {\scriptsize\textcolor{better}{-7.5}} & 50.8 {\scriptsize\textcolor{better}{+15.0}} & 3.0 {\scriptsize\textcolor{worse}{+14.2}}  \\
Qwen2.5-VL-32B    & 38.9 {\scriptsize\textcolor{better}{-7.4}} & 41.5 {\scriptsize\textcolor{better}{+30.2}} & \cellcolor{first}0.0 {\scriptsize\textcolor{better}{+0}} & 42.6 {\scriptsize\textcolor{better}{-11.6}} & 30.0 {\scriptsize\textcolor{better}{+20.8}} & \cellcolor{first}0.0 {\scriptsize\textcolor{worse}{+10.9}}  & 39.7 {\scriptsize\textcolor{better}{-8.3}} & 39.2 {\scriptsize\textcolor{better}{+28.0}} & \cellcolor{first}0.0 {\scriptsize\textcolor{better}{+0}} & 40.8 {\scriptsize\textcolor{better}{-9.5}} & 35.8 {\scriptsize\textcolor{better}{+25.5}} & \cellcolor{first}0.0 {\scriptsize\textcolor{worse}{+5.5}}  \\
Qwen2.5-VL-7B     & 34.0 {\scriptsize\textcolor{worse}{+10.8}} & 33.7 {\scriptsize\textcolor{better}{+6.3}} & 28.3 {\scriptsize\textcolor{worse}{+23.9}}  & 39.1 {\scriptsize\textcolor{worse}{+2.9}} & 20.5 {\scriptsize\textcolor{better}{+18.7}} & \cellcolor{first}0.0 {\scriptsize\textcolor{worse}{+20.1}}  & 35.0 {\scriptsize\textcolor{worse}{+9.2}} & 31.0 {\scriptsize\textcolor{better}{+9.6}} & 22.5 {\scriptsize\textcolor{worse}{+22.8}}  & 36.5 {\scriptsize\textcolor{worse}{+6.8}} & 27.1 {\scriptsize\textcolor{better}{+12.5}} & 14.2 {\scriptsize\textcolor{worse}{+22.0}}  \\
Qwen2.5-VL-3B     & 32.2 {\scriptsize\textcolor{worse}{+8.3}} & 32.8 {\scriptsize\textcolor{worse}{-1.9}} & \cellcolor{second}0.02 {\scriptsize\textcolor{worse}{+5.7}}  & 45.9 {\scriptsize\textcolor{better}{-2.7}} & 7.5 {\scriptsize\textcolor{better}{+8.5}}  & \cellcolor{first}0.0 {\scriptsize\textcolor{worse}{+17.1}}  & 35.0 {\scriptsize\textcolor{worse}{+6.1}} & 27.6 {\scriptsize\textcolor{better}{+0.6}} & \cellcolor{second}0.02 {\scriptsize\textcolor{worse}{+8.1}}  & 39.0 {\scriptsize\textcolor{worse}{+2.8}} & 20.2 {\scriptsize\textcolor{better}{+3.3}} & \cellcolor{second}0.01 {\scriptsize\textcolor{worse}{+11.4}}  \\
Qwen3-VL-32B      & \cellcolor{third}20.2 {\scriptsize\textcolor{worse}{+1.8}} & 80.7 {\scriptsize\textcolor{better}{+6.1}} & 0.3 {\scriptsize\textcolor{worse}{+0.1}}  & 25.1 {\scriptsize\textcolor{better}{-1.3}} & 51.9 {\scriptsize\textcolor{better}{+13.3}} & 7.4 {\scriptsize\textcolor{worse}{+0.0}}  & \cellcolor{third}21.2 {\scriptsize\textcolor{worse}{+1.1}} & 74.8 {\scriptsize\textcolor{better}{+7.6}} & 1.7 {\scriptsize\textcolor{worse}{+0.1}}  & 22.6 {\scriptsize\textcolor{worse}{+0.2}} & 66.3 {\scriptsize\textcolor{better}{+9.7}} & 3.9 {\scriptsize\textcolor{worse}{+0.0}}  \\
Qwen3-VL-8B       & 23.5 {\scriptsize\textcolor{worse}{+5.3}} & 67.0 {\scriptsize\textcolor{better}{+8.1}} & \cellcolor{first}0.0 {\scriptsize\textcolor{worse}{+1.3}}  & 35.3 {\scriptsize\textcolor{better}{-10.2}} & 34.2 {\scriptsize\textcolor{better}{+20.7}} & \cellcolor{third}3.4 {\scriptsize\textcolor{worse}{+15.3}}  & 25.9 {\scriptsize\textcolor{worse}{+2.2}} & 60.3 {\scriptsize\textcolor{better}{+11.0}} & 0.7 {\scriptsize\textcolor{worse}{+4.4}}  & 29.4 {\scriptsize\textcolor{better}{-2.4}} & 50.6 {\scriptsize\textcolor{better}{+14.4}} & 1.7 {\scriptsize\textcolor{worse}{+8.4}}  \\
Qwen3-VL-4B       & 30.6 {\scriptsize\textcolor{worse}{+1.4}} & 57.4 {\scriptsize\textcolor{better}{+10.3}} & \cellcolor{first}0.0 {\scriptsize\textcolor{worse}{+1.5}}  & 36.2 {\scriptsize\textcolor{better}{-7.0}} & 38.1 {\scriptsize\textcolor{better}{+6.4}} & \cellcolor{second}0.2 {\scriptsize\textcolor{worse}{+9.5}}  & 31.8 {\scriptsize\textcolor{better}{-0.3}} & 53.4 {\scriptsize\textcolor{better}{+9.5}} & \cellcolor{third}0.04 {\scriptsize\textcolor{worse}{+3.0}}  & 33.4 {\scriptsize\textcolor{better}{-2.8}} & 47.8 {\scriptsize\textcolor{better}{+8.4}} & \cellcolor{third}0.1 {\scriptsize\textcolor{worse}{+5.6}}  \\
Qwen3-VL-2B       & 64.5 {\scriptsize\textcolor{better}{-2.9}} & 32.1 {\scriptsize\textcolor{worse}{-7.6}} & 5.2 {\scriptsize\textcolor{worse}{+5.2}}  & 59.3 {\scriptsize\textcolor{better}{-8.6}} & NaN {\scriptsize\textcolor{better}{+24.1}}     & 12.5 {\scriptsize\textcolor{worse}{+0.2}}  & 63.4 {\scriptsize\textcolor{better}{-4.1}} & NaN     & 6.7 {\scriptsize\textcolor{worse}{+4.1}}  & 61.9 {\scriptsize\textcolor{better}{-5.7}} & NaN     & 8.9 {\scriptsize\textcolor{worse}{+2.7}}  \\
Intern3.5-VL-38B & 35.2 {\scriptsize\textcolor{worse}{+21.4}} & 56.7 {\scriptsize\textcolor{worse}{-31.0}} & 37.1 {\scriptsize\textcolor{better}{-36.3}} & 26.5 {\scriptsize\textcolor{worse}{+4.2}} & 23.3 {\scriptsize\textcolor{better}{+26.4}} & 43.1 {\scriptsize\textcolor{better}{-28.5}} & 33.4 {\scriptsize\textcolor{worse}{+17.9}} & 49.9 {\scriptsize\textcolor{worse}{-22.4}} & 38.3 {\scriptsize\textcolor{better}{-34.6}} & 30.8 {\scriptsize\textcolor{worse}{+12.8}} & 40.0 {\scriptsize\textcolor{worse}{-2.3}} & 40.1 {\scriptsize\textcolor{better}{-32.4}} \\
Intern3.5-VL-14B & 65.2 {\scriptsize\textcolor{better}{-4.7}} & -22.3 {\scriptsize\textcolor{better}{+42.8}} & 0.2 {\scriptsize\textcolor{worse}{+0.0}}  & 39.5 {\scriptsize\textcolor{better}{-5.0}} & 10.3 {\scriptsize\textcolor{better}{+16.4}} & 6.8 {\scriptsize\textcolor{worse}{+0.0}} & 60.0 {\scriptsize\textcolor{better}{-4.7}} & -15.6 {\scriptsize\textcolor{better}{+36.3}} & 1.5 {\scriptsize\textcolor{worse}{+0.0}} & 52.3 {\scriptsize\textcolor{better}{-4.8}} & -6.0 {\scriptsize\textcolor{better}{+29.6}} & 3.5 {\scriptsize\textcolor{worse}{+0.0}} \\
Intern3.5-VL-4B  & 43.7 {\scriptsize\textcolor{worse}{+12.0}} & 18.0 {\scriptsize\textcolor{worse}{-6.8}} & 0.3 {\scriptsize\textcolor{worse}{+0.0}}  & 37.0 {\scriptsize\textcolor{better}{-1.0}} & 5.6 {\scriptsize\textcolor{better}{+10.9}}  & 10.9 {\scriptsize\textcolor{worse}{+0.4}} & 42.3 {\scriptsize\textcolor{worse}{+9.3}} & 15.5 {\scriptsize\textcolor{worse}{-3.3}} & 2.5 {\scriptsize\textcolor{worse}{+0.1}} & 40.4 {\scriptsize\textcolor{worse}{+5.5}} & 11.8 {\scriptsize\textcolor{better}{+2.1}} & 5.6 {\scriptsize\textcolor{worse}{+0.2}} \\
\midrule
ProgressLM-3B-SFT & 19.0 & 72.4 & 6.5  & 29.1 & 46.3 & 9.2  & 21.1 & 67.0 & 7.0  & 24.0 & 59.3 & 7.8  \\
ProgressLM-3B-RL  & \cellcolor{first}13.8 & \cellcolor{first}90.1 & 8.5  & \cellcolor{third}21.2 & \cellcolor{first}63.9 & 5.6  & \cellcolor{first}15.3 & \cellcolor{first}84.8 & 7.9  & \cellcolor{first}17.5 & \cellcolor{first}77.0 & 7.0  \\
\bottomrule
\end{tabular}
}
\end{table*}

\section{Evaluation on \textsc{Progress-Bench}}

\paragraph{Experimental Setup.}
We evaluate 14 VLMs (2B–72B), including GPT-5/mini~\citep{openai2024gpt5systemcard}, Qwen~\citep{bai2025qwen2}, and Intern models~\citep{wang2025internvl3}, on \textsc{Progress-Bench}.
Each model is tested with \emph{direct prediction}, \emph{training-free reasoning}, or \emph{training-based reasoning} (\textsc{ProgressLM-SFT/RL} based on Qwen2.5-VL-3B).
All training tasks are disjoint from the benchmark to ensure generalization.

\paragraph{Evaluation Design.}
We evaluate progress estimation using four metrics that capture pointwise accuracy, temporal consistency, and answerability awareness.
\textbf{Normalized Score Error (NSE)} measures pointwise error on answerable samples.
\textbf{Progress Rank Correlation (PRC)} computes the Spearman rank correlation between predicted and ground-truth progress sequences along each trajectory, averaged across trajectories, with higher values close to $1$ indicating better temporal ordering~\citep{spearman1961proof}.
\textbf{Answerable False Rejection Rate (AFRR)} measures the fraction of answerable samples predicted as unanswerable, while
\textbf{Unanswerable Detection Accuracy (UDA)} measures the fraction of unanswerable samples correctly identified.

\begin{table*}[t]
\centering
\caption{
Same vs. cross-view performance on vision-based demonstrations.
\colorbox{first}{Best}, \colorbox{second}{Second}, and \colorbox{third}{Third} results are highlighted; colored deltas indicate the effect of training-free reasoning (\textcolor{better}{better} or \textcolor{worse}{worse}).
}
\label{tab:same_cross_view}
\resizebox{\textwidth}{!}{%
\begin{tabular}{lccccccccc}
\toprule
\textbf{Model} &
\multicolumn{3}{c}{\textbf{Same-View}} &
\multicolumn{3}{c}{\textbf{Cross-View}} &
\multicolumn{3}{c}{\textbf{Delta (Same $\rightarrow$ Cross)}} \\
\cmidrule(lr){2-4} \cmidrule(lr){5-7} \cmidrule(lr){8-10}
& NSE$\downarrow$ & PRC$\uparrow$ & AFRR$\downarrow$ & NSE$\downarrow$ & PRC$\uparrow$ & AFRR$\downarrow$ & $\Delta$NSE$\downarrow$ & $\Delta$PRC$\uparrow$ & $\Delta$AFRR$\downarrow$ \\
\midrule
GPT-5            & \cellcolor{second}14.6 {\scriptsize\textcolor{better}{-0.4}} & \cellcolor{third}89.4 {\scriptsize\textcolor{better}{+1.0}} & \cellcolor{first}0.0 {\scriptsize\textcolor{better}{+0}} & \cellcolor{third}20.5 {\scriptsize\textcolor{better}{-0.6}} & \cellcolor{second}89.4 {\scriptsize\textcolor{worse}{-0.6}} & 1.8 {\scriptsize\textcolor{better}{-0.3}} & +5.9 {\scriptsize\textcolor{better}{-0.2}} & 0.0 {\scriptsize\textcolor{worse}{-1.6}} & +1.8 {\scriptsize\textcolor{better}{-0.3}} \\
GPT-5-mini       & 19.8 {\scriptsize\textcolor{worse}{+0.8}} & 84.1 {\scriptsize\textcolor{worse}{-0.6}} & \cellcolor{third}0.4 {\scriptsize\textcolor{better}{-0.4}}  & 21.1 {\scriptsize\textcolor{worse}{+0.6}} & \cellcolor{first}89.1 {\scriptsize\textcolor{worse}{-0.6}} & \cellcolor{third}0.4 {\scriptsize\textcolor{better}{-0.2}} & +1.3 {\scriptsize\textcolor{better}{-0.2}} & +5.0 {\scriptsize{+0.0}} & 0.0 {\scriptsize\textcolor{worse}{+0.2}} \\
Qwen2.5-VL-72B    & 16.8 {\scriptsize\textcolor{worse}{+5.0}} & 83.9 {\scriptsize\textcolor{worse}{-5.0}} & 0.4 {\scriptsize\textcolor{better}{-0.5}}  & 30.9 {\scriptsize\textcolor{better}{-5.4}} & 50.7 {\scriptsize\textcolor{better}{+26.4}} & 13.4 {\scriptsize\textcolor{better}{-7.7}} & +14.1 {\scriptsize\textcolor{better}{-10.4}} & -33.2 {\scriptsize\textcolor{better}{+31.4}} & +13.0 {\scriptsize\textcolor{better}{-7.2}} \\
Qwen2.5-VL-32B    & 21.4 {\scriptsize\textcolor{worse}{+6.2}} & 72.9 {\scriptsize\textcolor{better}{+0.1}} & \cellcolor{first}0.0 {\scriptsize\textcolor{better}{+0}} & 45.7 {\scriptsize\textcolor{better}{-12.7}} & 29.4 {\scriptsize\textcolor{better}{+40.5}} & \cellcolor{first}0.0 {\scriptsize\textcolor{better}{+0}} & +24.3 {\scriptsize\textcolor{better}{-18.9}} & -43.5 {\scriptsize\textcolor{better}{+40.4}} & 0.0 {\scriptsize\textcolor{worse}{+0.0}} \\
Qwen2.5-VL-7B     & 27.3 {\scriptsize\textcolor{worse}{+11.1}} & 51.9 {\scriptsize\textcolor{worse}{-7.6}} & 30.6 {\scriptsize\textcolor{worse}{+26.5}}  & 36.5 {\scriptsize\textcolor{worse}{+10.6}} & 26.7 {\scriptsize\textcolor{better}{+3.7}} & 26.0 {\scriptsize\textcolor{worse}{+21.4}} & +9.2 {\scriptsize\textcolor{better}{-0.5}} & -25.2 {\scriptsize\textcolor{better}{+11.3}} & -4.6 {\scriptsize\textcolor{better}{-5.1}} \\
Qwen2.5-VL-3B     & 29.2 {\scriptsize\textcolor{worse}{+9.5}} & 43.0 {\scriptsize\textcolor{worse}{-11.1}} & 9.9 {\scriptsize\textcolor{worse}{+5.8}}  & 33.4 {\scriptsize\textcolor{worse}{+7.9}} & 28.9 {\scriptsize\textcolor{worse}{-2.5}} & 6.5 {\scriptsize\textcolor{worse}{+4.4}} & +4.2 {\scriptsize\textcolor{better}{-1.6}} & -14.1 {\scriptsize\textcolor{better}{+8.6}} & -3.4 {\scriptsize\textcolor{better}{-1.4}} \\
Qwen3-VL-32B      & \cellcolor{third}15.9 {\scriptsize\textcolor{worse}{+2.6}} & \cellcolor{second}88.3 {\scriptsize\textcolor{worse}{-8.0}} & \cellcolor{first}0.0 {\scriptsize\textcolor{better}{+0}} & \cellcolor{second}21.9 {\scriptsize\textcolor{worse}{+1.4}} & \cellcolor{third}77.8 {\scriptsize\textcolor{better}{+11.0}} & 0.4 {\scriptsize\textcolor{better}{-3.5}} & +6.0 {\scriptsize\textcolor{better}{-1.2}} & -10.5 {\scriptsize\textcolor{better}{+19.0}} & +0.4 {\scriptsize\textcolor{better}{-3.5}} \\
Qwen3-VL-8B       & 19.1 {\scriptsize\textcolor{worse}{+5.6}} & 81.7 {\scriptsize\textcolor{worse}{-9.5}} & \cellcolor{first}0.0 {\scriptsize\textcolor{worse}{+1.3}}  & 25.2 {\scriptsize\textcolor{worse}{+5.2}} & 61.3 {\scriptsize\textcolor{better}{+16.0}} & \cellcolor{first}0.0 {\scriptsize\textcolor{worse}{+1.3}} & +6.1 {\scriptsize\textcolor{better}{-0.4}} & -20.4 {\scriptsize\textcolor{better}{+25.5}} & 0.0 {\scriptsize\textcolor{worse}{+0.0}} \\
Qwen3-VL-4B       & 21.6 {\scriptsize\textcolor{worse}{+3.5}} & 72.3 {\scriptsize\textcolor{worse}{-3.2}} & \cellcolor{first}0.0 {\scriptsize\textcolor{worse}{+1.1}}  & 34.1 {\scriptsize\textcolor{worse}{+0.4}} & 51.6 {\scriptsize\textcolor{better}{+17.6}} & \cellcolor{first}0.0 {\scriptsize\textcolor{worse}{+1.6}} & +12.5 {\scriptsize\textcolor{better}{-3.1}} & -20.7 {\scriptsize\textcolor{better}{+20.8}} & 0.0 {\scriptsize\textcolor{worse}{+0.5}} \\
Qwen3-VL-2B       & 60.0 {\scriptsize\textcolor{better}{-1.6}} & 35.0 {\scriptsize\textcolor{worse}{-4.6}} & \cellcolor{second}0.1 {\scriptsize\textcolor{worse}{+9.8}}  & 66.2 {\scriptsize\textcolor{better}{-3.4}} & 30.9 {\scriptsize\textcolor{worse}{-12.6}} & \cellcolor{first}0.0 {\scriptsize\textcolor{worse}{+3.5}} & +6.2 {\scriptsize\textcolor{better}{-1.8}} & -4.1 {\scriptsize\textcolor{worse}{-8.0}} & -0.1 {\scriptsize\textcolor{better}{-6.3}} \\
Intern3.5-VL-38B & 27.6 {\scriptsize\textcolor{worse}{+28.4}} & 74.8 {\scriptsize\textcolor{worse}{-51.5}} & 0.5 {\scriptsize\textcolor{worse}{+0.1}} & 38.1 {\scriptsize\textcolor{worse}{+18.6}} & 49.7 {\scriptsize\textcolor{worse}{-17.9}} & 51.3 {\scriptsize\textcolor{better}{-50.4}} & +10.5 {\scriptsize\textcolor{better}{-9.8}} & -25.1 {\scriptsize\textcolor{better}{+33.6}} & +50.8 {\scriptsize\textcolor{better}{-50.5}} \\
Intern3.5-VL-14B & 62.1 {\scriptsize\textcolor{worse}{+0.0}} & 24.2 {\scriptsize\textcolor{worse}{-4.5}} & 0.0 {\scriptsize\textcolor{worse}{+0.0}} & 66.4 {\scriptsize\textcolor{better}{-6.4}} & -40.3 {\scriptsize\textcolor{better}{+65.4}} & \cellcolor{first}0.0 {\scriptsize\textcolor{better}{+0}} & +4.3 {\scriptsize\textcolor{better}{-6.4}} & -64.5 {\scriptsize\textcolor{better}{+69.9}} & 0.0 {\scriptsize\textcolor{worse}{+0.0}} \\
Intern3.5-VL-4B  & 44.4 {\scriptsize\textcolor{worse}{+9.6}} & 33.1 {\scriptsize\textcolor{worse}{-18.9}} & 0.7 {\scriptsize\textcolor{better}{-0.4}} & 43.5 {\scriptsize\textcolor{worse}{+12.8}} & 12.1 {\scriptsize\textcolor{worse}{-8.6}} & 0.2 {\scriptsize\textcolor{better}{-0.2}} & -0.9 {\scriptsize\textcolor{worse}{+3.2}} & -21.0 {\scriptsize\textcolor{better}{+10.3}} & -0.5 {\scriptsize\textcolor{worse}{+0.2}} \\
\midrule
ProgressLM-3B-SFT & 15.5 & 84.0 & 0.6  & 20.4 & 67.8 & 8.8 & +4.9 & -16.2 & +8.2 \\
ProgressLM-3B-RL  & \cellcolor{first}10.3 & \cellcolor{first}93.5 & \cellcolor{second}0.1  & \cellcolor{first}15.2 & \cellcolor{third}88.8 & 11.7 & +4.9 & -4.7 & +11.6 \\
\bottomrule
\end{tabular}
}
\end{table*}

\begin{figure*}[t]
    \centering
    \includegraphics[width=1\linewidth]{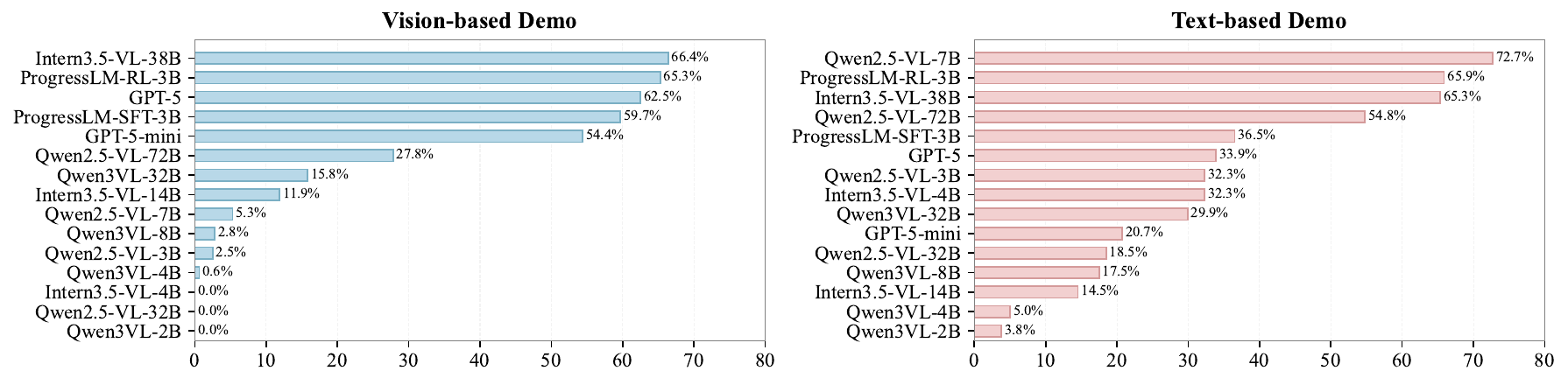}
    \caption{
    Unanswerable Detection Accuracy (UDA) across models under two settings.
    }
    \label{fig:nega-bar}
\end{figure*}

\subsection{Performance on Answerable Scenarios}

We first evaluate model performance on \textit{answerable} samples, where task progress is well-defined.
Table~\ref{tab:main_result} reports results under both vision-based and text-based demonstrations.

\paragraph{How well do current VLMs perform at progress estimation?}
\textbf{\textit{Overall, current VLMs show limited and highly unstable progress estimation under direct prediction.}}
Although strong models such as GPT-5 and Qwen2.5-VL-72B perform better than smaller counterparts, they remain highly sensitive to demonstration modality, with vision-based demonstrations consistently outperforming text-based ones.
We further observe abnormally low, negative, or undefined PRC values for several models, indicating collapsed or distorted progress rankings rather than meaningful ordinal reasoning.
As analyzed in Section~\ref{further_analysis_pred_score_distribution}, these failures stem from degenerate predicted score distributions, such as collapse to extreme or discrete values.
Finally, some models (e.g., Intern3.5-VL-38B) exhibit extremely high AFRR even on answerable samples, reflecting overly conservative rejection behavior rather than calibrated uncertainty.

\paragraph{Does training-free progress reasoning help?}
\textbf{\textit{Training-free reasoning yields conditional benefits and depends strongly on model capacity.}}
Large models (e.g., GPT-5 and Qwen2.5-VL-72B/32B) gain moderate improvements, primarily in PRC and occasionally in NSE.
In contrast, smaller models often see marginal or negative effects, including increased NSE or AFRR, suggesting that limited-capacity models may follow the reasoning format without genuinely improving progress understanding.

\paragraph{Does training-based progress reasoning help?}
\textbf{\textit{Explicit training enables robust progress reasoning even at small scale.}}
\textsc{ProgressLM} consistently improves over the base 3B model across all answerable metrics, with \textsc{ProgressLM-RL-3B} achieving the strongest macro-averaged NSE and PRC among all evaluated models.
These gains demonstrate that effective progress reasoning is not solely driven by scale, but can be reliably learned through targeted supervision and optimization.

\subsection{Robustness to Viewpoint Changes}

To assess robustness to viewpoint changes, we decompose vision-based results into \textit{same-view} and \textit{cross-view} settings (Table~\ref{tab:same_cross_view}).

\paragraph{How do current VLMs handle viewpoint changes?}
\textbf{\textit{Most VLMs degrade substantially under cross-view observations.}}
Across models, cross-view settings yield higher NSE and lower PRC than same-view ones, with especially large drops for small and medium-sized models.
This pattern suggests that many VLMs rely heavily on viewpoint-dependent visual similarity rather than viewpoint-invariant progress reasoning.

\paragraph{Does progress reasoning improve cross-view robustness?}
\textbf{\textit{Training-free progress reasoning provides limited and conditional gains.}}
Its effectiveness strongly depends on baseline capability: weaker models benefit little, while strong models (e.g., GPT-5) show only modest cross-view improvements, often at the expense of same-view performance.
When gains occur, they primarily appear in PRC rather than NSE, indicating improved temporal ordering rather than pointwise accuracy.
\textbf{\textit{In contrast, robust cross-view generalization emerges only through explicit training.}}
\textsc{ProgressLM-3B-RL} consistently exhibits smaller same-view to cross-view gaps, demonstrating improved robustness beyond surface-level visual correspondence.

\subsection{Unanswerable Case Recognition}
\paragraph{Can models recognize when progress estimation is not possible?}
\textit{\textbf{Most VLMs fail to reliably recognize unanswerable cases.}}
As shown in Figure~\ref{fig:nega-bar}, most models still produce progress scores even when the input is inherently ambiguous, indicating limited awareness of unanswerability.
\textbf{\textit{In contrast, \textsc{ProgressLM} consistently identifies unanswerable cases under both vision-based and text-based demonstrations, avoiding forced predictions.}}
This behavior is further strengthened by reinforcement learning, with \textsc{ProgressLM-3B-RL} achieving the highest or near-highest unanswerable recognition accuracy.
However, strong unanswerable detection alone is insufficient.
For example, \textsc{InternVL-3.5-38B} attains relatively high UDA but exhibits an extremely high Answerable False Rejection Rate (AFRR) (Table~\ref{tab:main_result}), indicating overly conservative behavior.

\section{Further Analysis}

\subsection{Predicted Score Distribution}
\label{further_analysis_pred_score_distribution}
\begin{figure}[t]
    \centering
    \includegraphics[width=1\linewidth]{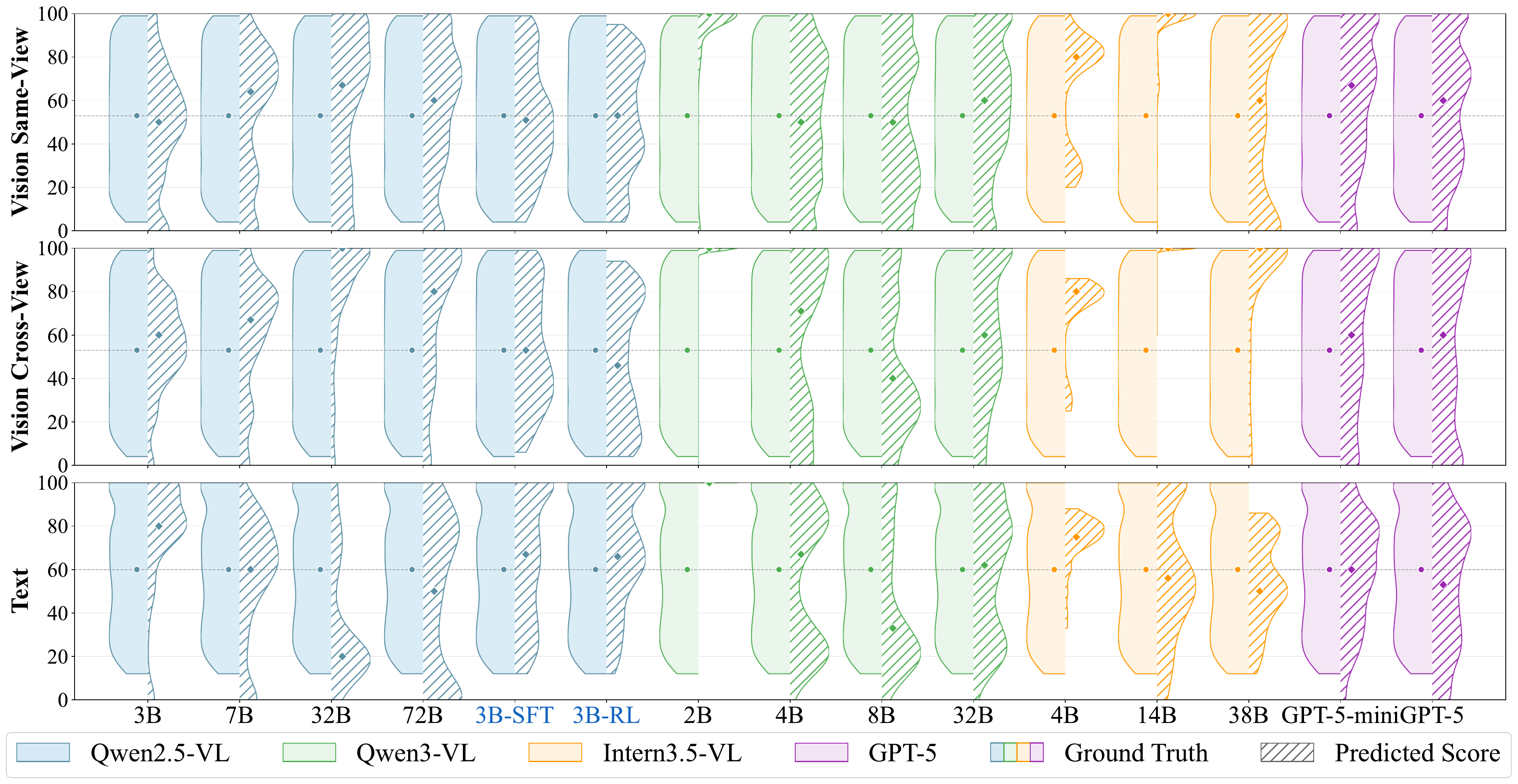}
    \caption{
Distribution of predicted progress scores.
}
\label{fig:predict_score_distribution}
\end{figure}
\paragraph{What patterns emerge in predicted progress score distributions?}
Predicted scores often cluster around specific values rather than varying smoothly.
We observe four recurring patterns in Figure~\ref{fig:predict_score_distribution}:
\textbf{\textit{(i) single-peak collapse}} at extreme values (e.g., $0\%$ or $100\%$);
\textbf{\textit{(ii) multi-peak clustering}} around a few heuristic anchors;
\textbf{\textit{(iii) central-peaked distributions}} concentrated near $\sim50\%$, reflecting default or uncertain responses;
and \textbf{\textit{(iv) smooth continuous distributions}} spanning $[0,100\%]$, indicating sensitivity to intermediate task states.
Notably, \textsc{ProgressLM-3B-SFT} and \textsc{ProgressLM-3B-RL} consistently exhibit the fourth pattern, whereas most base VLMs fall into the first three.
These distributional failures directly explain unstable or distorted rank correlations, underscoring that fine-grained progress reasoning emerges only through explicit learning.

\begin{figure}[t]
    \centering
    \includegraphics[width=1\linewidth]{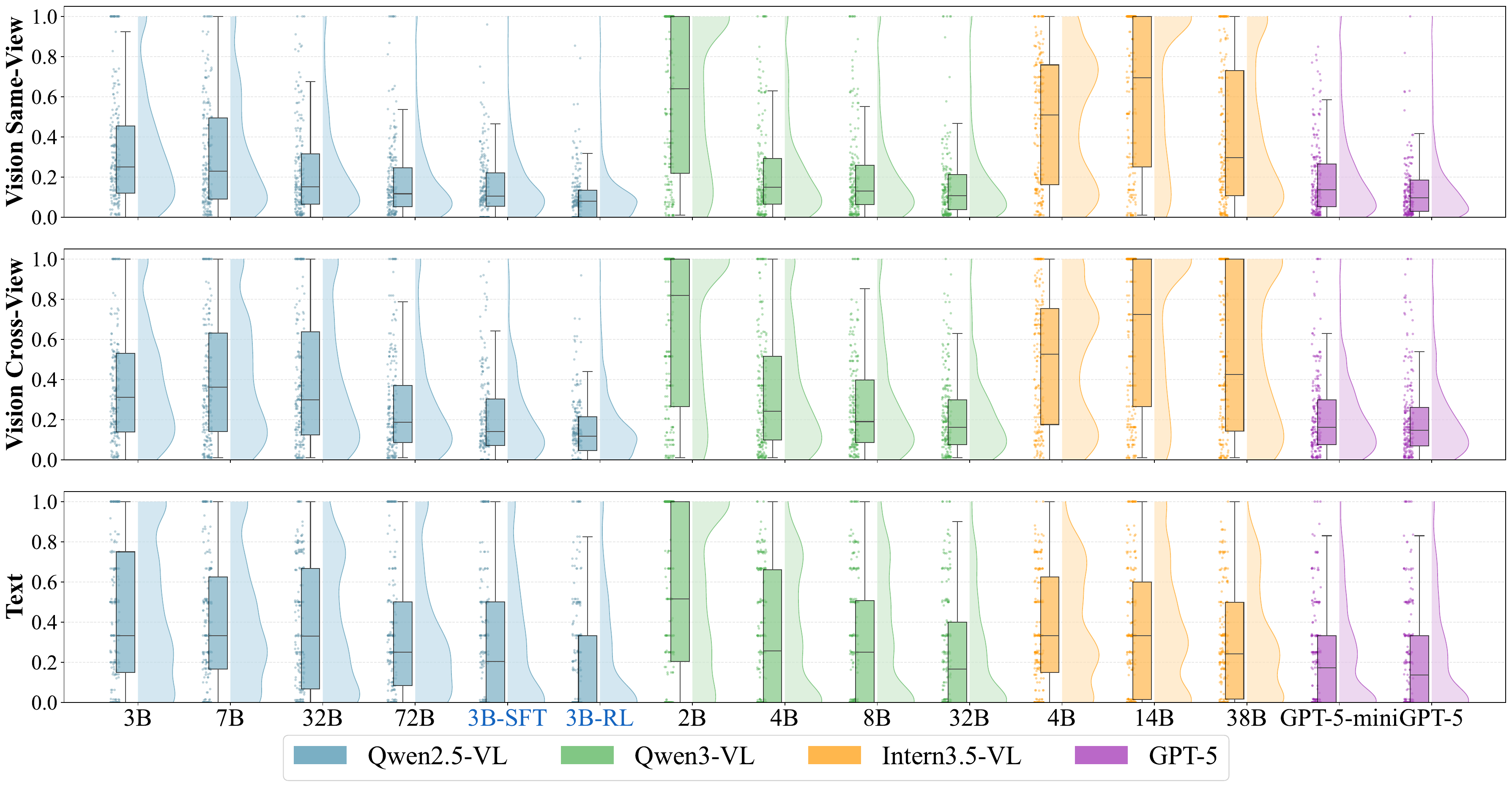}
    \caption{
        Raincloud plots of per-sample normalized score prediction error across models.
    }    \label{fig:per_sample_error_raincloud}
\end{figure}

\subsection{Per-sample Error Distribution}

\paragraph{What do per-sample error distributions reveal?}
\textbf{\textit{Smaller models exhibit unstable errors, while explicit progress learning substantially improves robustness.}}
In Figure~\ref{fig:per_sample_error_raincloud}, smaller models show broad, heavy-tailed error distributions, reflecting frequent large deviations and unstable progress estimates, consistent with their collapsed or spiky score distributions.
Larger models produce more concentrated errors closer to zero.
Notably, both \textsc{ProgressLM-3B-SFT} and \textsc{ProgressLM-3B-RL} markedly tighten the error distribution relative to the base 3B model, with \textsc{ProgressLM-3B-RL} especially suppressing extreme-error cases.
These results indicate that explicitly learned progress reasoning improves not only average accuracy but also per-sample robustness.

\begin{figure}[t]
    \centering
    \includegraphics[width=\linewidth]{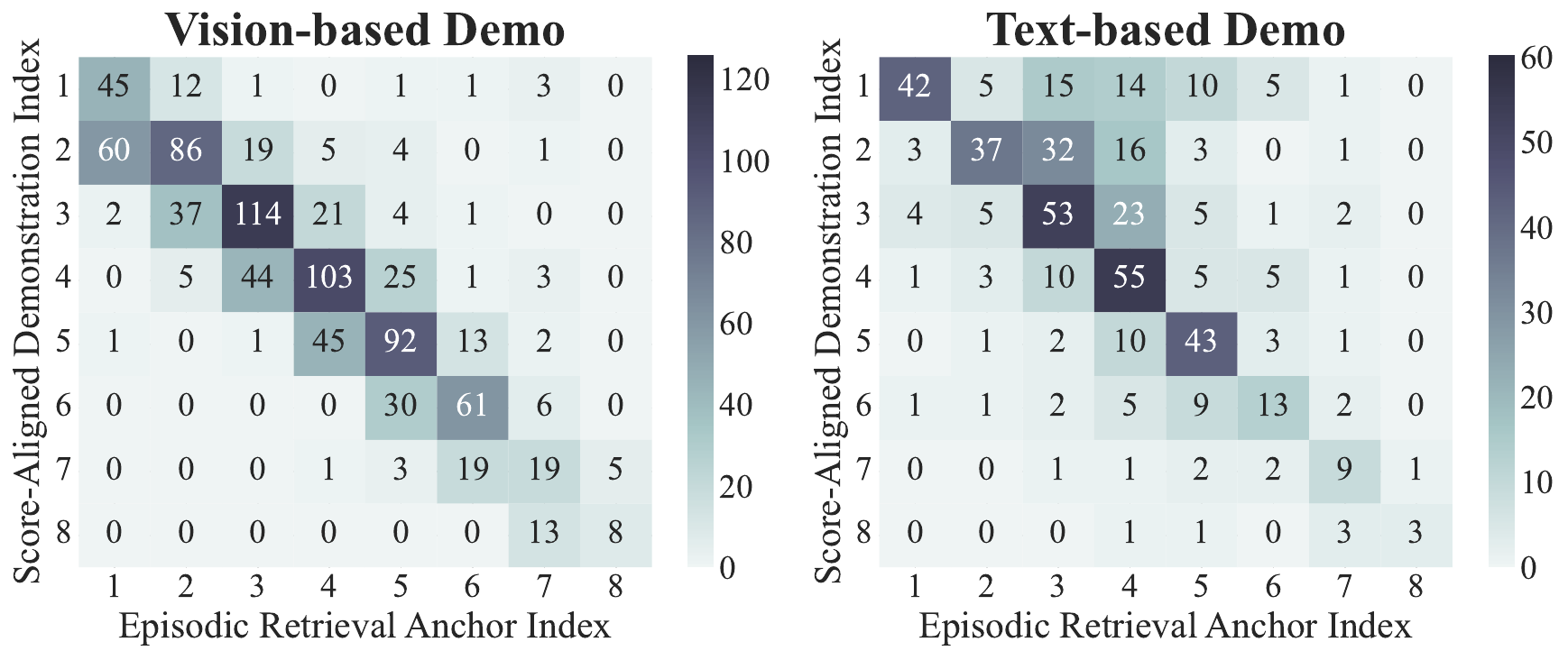}
    \vspace{-1.0em}
    \caption{
Coupling between the two stages progress reasoning of \textsc{ProgressLM}.
}
    \vspace{-1em}
    \label{fig:ref-score-corr}
\end{figure}

\subsection{Coupled Progress Reasoning}

\paragraph{Are the two reasoning stages truly coupled?}
\textbf{\textit{Yes—the first-stage anchor directly constrains second-stage progress estimation.}}
Figure~\ref{fig:ref-score-corr} analyzes the relationship between the \textit{Episodic Retrieval Anchor Index} (the step retrieved in episodic retrieval) and the \textit{Score-Aligned Demonstration Index} (the step whose annotated progress best matches the predicted score).
If the two stages are coupled, these indices should align, yielding a near-diagonal pattern.
The strong diagonal concentration observed for our training-based model confirms that anchor retrieval is not auxiliary, but actively guides fine-grained progress estimation via second-stage \textit{mental simulation}.

\begin{figure}[t]
    \centering
    \includegraphics[width=\linewidth]{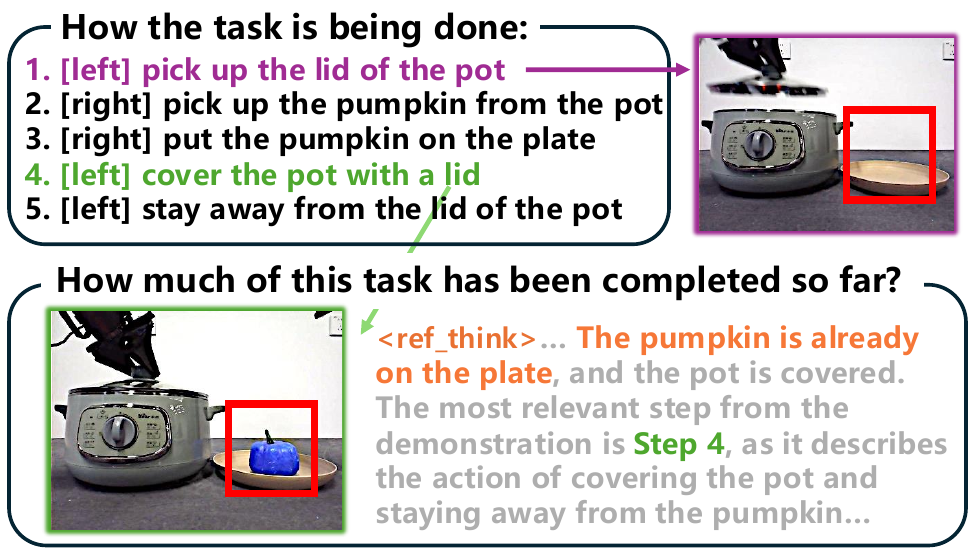}
    \vspace{-1.0em}
    \caption{
Illustration of implicit state accumulation required by text-based demonstrations.
}
\vspace{-1em}
    \label{fig:visual-case}
\end{figure}

\subsection{Implicit State Accumulation}
\label{further_analysis_text_hard}

\paragraph{Why are text-based demonstrations harder than vision-based ones?}
\textit{\textbf{Text-based demonstrations require implicit state accumulation rather than direct state matching.}}
Unlike vision-based demonstrations that explicitly show complete world states, text-based demonstrations describe actions whose effects must be integrated over time.
As illustrated in Figure~\ref{fig:visual-case}, Step~1 and Step~4 both involve interacting with the pot lid, yet differ in an implicit state variable: whether the pumpkin has already been removed and placed on the plate.
Disambiguating these steps requires integrating the intervening actions (Steps~2 and~3), highlighting that text-based progress estimation depends on tracking latent state evolution rather than surface-level action semantics.

\begin{table}[t]
\centering
\small
\caption{Results on larger Qwen2.5-VL.}
\label{tab:scale_7b}
\begin{tabular}{lccc}
\toprule
\multicolumn{4}{c}{\textbf{Vision-based Demo}} \\
\midrule
Model & PRC$\uparrow$ & NSE$\downarrow$ & AFRR$\downarrow$ \\
\midrule
No think & 33.7 & 34.0 & 28.3 \\
Training-free & 40.0 & 44.8 & 52.2 \\
Training-based & \textbf{85.7} & \textbf{13.4} & \textbf{32.4} \\
\midrule
\multicolumn{4}{c}{\textbf{Text-based Demo}} \\
\midrule
No think & 20.5 & 39.1 & \textbf{0.0} \\
Training-free & 39.2 & 42.0 & 20.1 \\
Training-based & \textbf{50.5} & \textbf{26.6} & 1.4 \\
\bottomrule
\end{tabular}
\end{table}

\subsection{Generalization Across Model Scales.}
We further evaluate scalability on a 7B model. 
As shown in Table~\ref{tab:scale_7b}, the training-based approach consistently outperforms both no-thinking and training-free baselines across metrics. 
Notably, despite training the RL stage for only one epoch due to computational constraints, the 7B model already exhibits substantial gains, especially in the vision-based setting. 
These results suggest that the proposed supervision effectively scales with model size.

\section{Related Work}

\paragraph{Progress Estimation.}
Early progress estimation methods predominantly rely on task-specific or expert models trained within fixed tasks or environments~\citep{yang2024rank2reward, chen2025sarmstageawarerewardmodeling, ma2024dreureka}, which limits their ability to generalize beyond the training distribution.
Some approaches estimate progress by measuring distances in latent feature space~\citep{ma2022vip, ma2023liv, escontrela2023video, lee2021generalizable}, but they struggle to model fine-grained intermediate progress.
More recently, VLMs have been applied to progress estimation, but typically through indirect formulations~\citep{alakuijala2025video, ma2024vision, zhai2025vision}.
For example, progress is inferred via trajectory reordering by shuffling frames~\citep{ma2024vision}, or by aggregating pairwise relative progress comparisons~\citep{zhai2025vision}.
In these settings, progress estimates are not independent, but strongly coupled to other predictions within the same trajectory, causing each estimate to depend on the entire sequence context.
In contrast, humans can infer task progress from a single observation by reasoning over underlying task dynamics, highlighting progress estimation as a long-horizon and dynamic reasoning problem.

\paragraph{Progress Reasoning in VLMs.}
Recent advances in VLMs have substantially improved static visual reasoning capabilities from single or multiple images~\citep{zhang2025vlm2, lee2025multiverse, yuan2025autodrive,yuan2025video,yuan2026if, pi2024image,xia2025sportr,wu2026camreasoner,wu2025refineshot,xing2025re,ge2025framemind,ge2025mrfd,qian2025decalign,qian2025dyncim,han2025climd, jia2024leopard,wang2026streetdesignaimultipersonaevaluationinclusive,wang2026visionlanguagemodelsunderstand,dai2026personaawareexplainablebikeabilityassessment,du2026multimodalgenerativeengineoptimization,nian2025jaildamjailbreakdetectionadaptive}, largely focusing on snapshot-level perception.
However, dynamic reasoning under partial observation requires models to reason over long-horizon task evolution, infer latent state transitions, and maintain an implicit world model~\citep{yin2025spatial, wang2025vagen}.
Progress reasoning naturally embodies these challenges, as it demands fine-grained understanding of intermediate task states from a single observation.
Nevertheless, most existing VLM-based approaches focus on \emph{coarse task completion judgments}, often reducing progress estimation to binary decisions and overlooking whether models can reason about \emph{intermediate progress from a single observation}~\citep{fan2022minedojo, cui2022can, wang2024rl, lu2025robofac, duan2024aha, luo2025roboreflect, dai2025racer}.
This gap motivates our work, which studies progress estimation as a structured reasoning capability in VLMs rather than a heuristic or static prediction.

\section{Conclusion}

We study progress estimation as a long-horizon, dynamic reasoning problem beyond static visual understanding.
We introduce \textsc{Progress-Bench} to systematically evaluate progress reasoning from a single observation under controlled variations of modality, viewpoint, and answerability.
Experiments on 14 VLMs show that existing models struggle with this task, exhibiting strong sensitivity to modality and viewpoint changes, degenerate progress predictions, and weak handling of unanswerable cases.
Our analyses expose systematic failure modes in existing VLMs and show that robust progress estimation emerges only when coarse anchor retrieval and fine-grained reasoning are explicitly learned.

\section{Limitations}
While this work provides a systematic study of progress reasoning in VLMs, it has several limitations.
\textsc{Progress-Bench} focuses on robotic manipulation tasks with relatively clear and monotonic progress, which may limit generalization to more open-ended scenarios with ambiguous goals or non-monotonic dynamics.
In addition, \textsc{ProgressLM} is trained on curated manipulation data with similar structural properties, and extending to substantially different task families may require further data or adaptation.

Despite these limitations, progress reasoning enables several practical applications.
It can serve as an anomaly detector for long-horizon systems by identifying stagnating or inconsistent progress, and as an online reward signal for reinforcement learning, providing dense and interpretable feedback.
It can also act as a data engine for training task-specific reward models, improving efficiency and reducing latency.

More broadly, the paradigm extends beyond embodied tasks to general agents (e.g., web agents)~\citep{wang2025explore, hu2025webcot}, where progress signals can support inference-time scaling via iterative refinement~\citep{fang2025webevolver, li2026verified, he2025openwebvoyager}, and potentially enable self-improving behaviors.
\section{Acknowledgments}
This work was supported in part by the NSF SKAI Institute and the Mellon Foundation.
\bibliography{custom}

\clearpage

\appendix
\section*{Appendix}
\label{sec:appendix}

\noindent
\begin{minipage}{\columnwidth} 
  \setlength{\cftbeforesecskip}{0.5em}
  \cftsetindents{section}{0em}{1.8em}
  \cftsetindents{subsection}{1em}{2.5em}
  \etoctoccontentsline{part}{Appendix}{
    \etocsettocstyle{}{}
    \localtableofcontents
  }
\end{minipage}

\section{Data Construction Details}

\subsection{Text Unanswerable Data}

We construct text negative samples by replacing key objects in task instructions to create misalignment between the visual observation and textual description. Given the task goal $G$, step-by-step instructions $\{s_1, s_2, \ldots, s_n\}$, and optionally a reference image $I$, we employ Qwen2.5-VL-72B \citep{bai2025qwen2} to generate modified instructions where the main manipulated object is replaced with a different object. The model is instructed to:
\begin{enumerate}
    \item Analyze the input state-to-estimate image and identify objects that could serve as plausible replacements
    \item Replace the target object in both the task goal and all step-by-step instructions
    \item Preserve the original sentence structure, action verbs, and spatial markers (\emph{e.g.}, \texttt{[left]}, \texttt{[right]}, \texttt{[towards]})
\end{enumerate}

The model outputs the modified task goal and instructions in a structured XML format, as shown in Table~\ref{prompt:text_nega}. This approach ensures that the edited instructions remain grammatically coherent and physically plausible while creating a clear mismatch between the visual scene and the textual description.

\begin{figure*}[t]
    \centering
    \includegraphics[width=\linewidth]{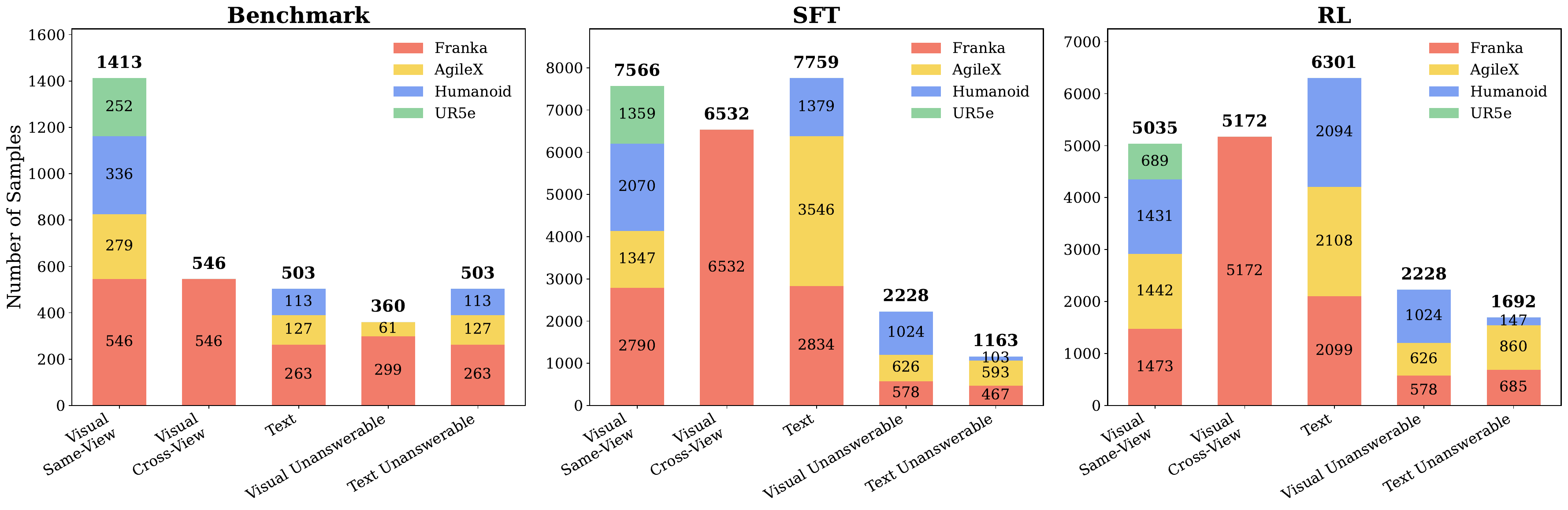}
    \caption{\textbf{Data distribution statistics across Benchmark, SFT, and RL splits.} This figure shows the distribution of samples produced by our data construction pipeline across Benchmark, Supervised Fine-Tuning (SFT), and Reinforcement Learning (RL) stages. Samples are organized by demonstration–observation setting (Visual Same-View, Visual Cross-View, Text, Visual Unanswerable, Text Unanswerable), with stacked bars denoting different robot platforms. Our constructed dataset spans four heterogeneous robotic platforms, including single-arm robot (Franka Emika Panda, UR5e), dual-arm robot (AgileX Cobot Magic V2.0), and humanoid robot (X-Humanoid Tien Kung), enabling evaluation and training across diverse embodiments.}
    \label{fig:data_barchart}
\end{figure*}

\begin{figure*}[ht]
    \centering
    \includegraphics[width=\linewidth]{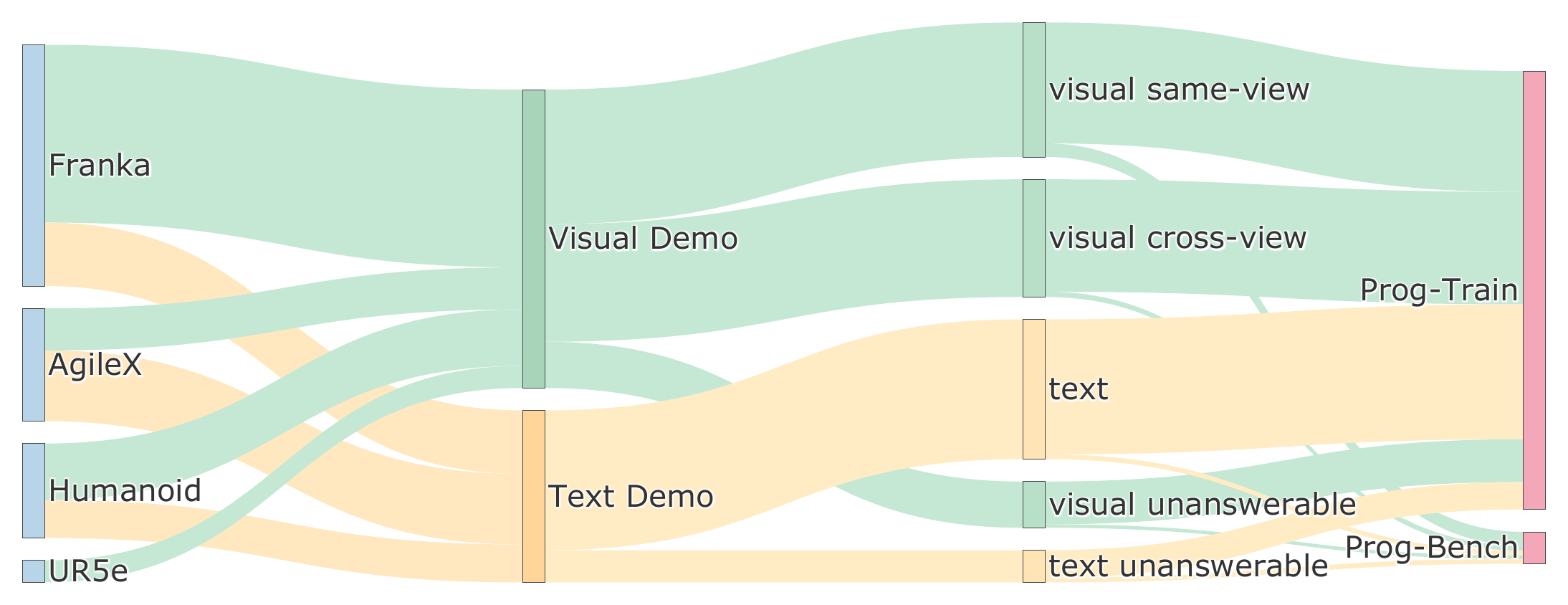}
    \caption{\textbf{Overview of the data construction pipeline.} This Sankey diagram illustrates how raw manipulation trajectories from four heterogeneous robotic platforms (Franka, AgileX, Humanoid, and UR5e) are transformed through our data construction process. Demonstrations are first organized into visual and text modalities, then further expanded into multiple demonstration–observation settings, including visual same-view, visual cross-view, text, as well as visual and text unanswerable cases. The resulting data are finally allocated to \textsc{\textbf{ProgressLM-45K}} for model training and \textsc{\textbf{ProgressLM-Bench}} for evaluation, highlighting the unified yet diversified pipeline that supports generalizable progress reasoning across embodiments, modalities, and answerability conditions.}
    \label{fig:app_sankey}
\end{figure*}

\begin{figure*}[ht]
    \centering
    \includegraphics[width=\linewidth]{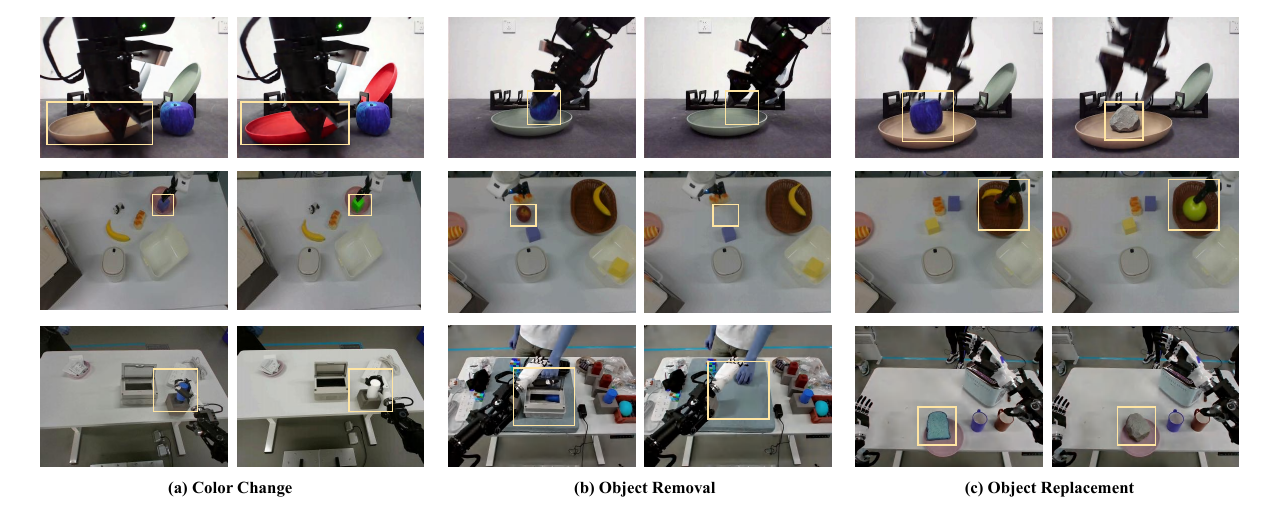}
    \caption{\textbf{Case Visualization of visual unanswerable samples construction via Image Editing}. To test whether models can recognize ill-defined progress, we construct visual unanswerable samples by breaking the semantic consistency between demonstrations and observations while preserving realism. Given an image at a specific manipulation step, we edit the key object using three strategies: (a) Color Change, altering object appearance; (b) Object Removal, eliminating the critical object; and (c) Object Replacement, substituting it with an incompatible one. As shown across diverse manipulation scenarios, these edits invalidate progress estimation and require models to correctly output N/A rather than relying on superficial visual matching.
    }
    \label{fig:data_barchart}
\end{figure*}

\subsection{Vision-Based Unanswerable Data} 

We construct visual unanswerable samples through a three-stage pipeline to evaluate model robustness against adversarial visual perturbations.

\noindent\textbf{Stage 1: Edit Prompt Generation.}
Given an input image $I$ along with the task goal $G$ and step-by-step instructions $\{s_1, s_2, \ldots, s_n\}$, we first identify the corresponding instruction $s_k$ that matches the current image state. We then employ Qwen2.5-VL-72B \citep{bai2025qwen2} to generate an editing prompt that would cause the instruction to be violated. The model is provided with a structured prompt containing: (1) the task goal and complete step-by-step instructions, (2) the input image to be edited, and (3) the specific instruction $s_k$ corresponding to the current image. The model is instructed to select one editing strategy from three predefined options:
\begin{enumerate}
    \item \textbf{Color Change}: Altering the color of critical objects (\emph{e.g.}, changing a red apple to green)
    \item \textbf{Object Replacement}: Replacing the target object with a semantically different object (\emph{e.g.}, replacing an egg with an orange)
    \item \textbf{Occlusion/Removal}: Hiding or removing key objects from the scene
\end{enumerate}

The model first reasons about which strategy would most effectively violate the instruction while maintaining visual realism, then outputs a concise editing prompt (maximum 20 words) in a structured XML format containing the reasoning process, selected strategy, and the final editing instruction. The complete prompt for is provided in Table~\ref{prompt:visual_nega}.

\noindent\textbf{Stage 2: Image Editing.}
We apply the generated editing prompts to the original images using Qwen-Image-Edit (20B)~\citep{wu2025qwenimagetechnicalreport}, a diffusion-based image editing model built upon the Diffusers library. The model takes the original image $I$ and the editing prompt $p$ as input, and generates an edited image $I'$ through an iterative denoising process. 

\noindent\textbf{Stage 3: Human Filtering.}
We develop a web-based annotation platform using \textbf{Gradio Platform} (As shown in Figure~\ref{fig:gradio}) to perform human quality control on the edited images. Annotators are presented with the edited image alongside the task goal, step-by-step instructions, editing strategy, and editing prompt. They assess whether the edit successfully violates the corresponding instruction while maintaining visual realism and naturalness. Each sample is labeled as either ``YES'' (keep) or ``NO'' (discard). Through this rigorous filtering process, we retain $23.5\%$ of the edited images that meet our quality criteria, resulting in a high-quality visual negative dataset.

\begin{figure*}[t]
    \centering
    \includegraphics[width=0.9\linewidth]{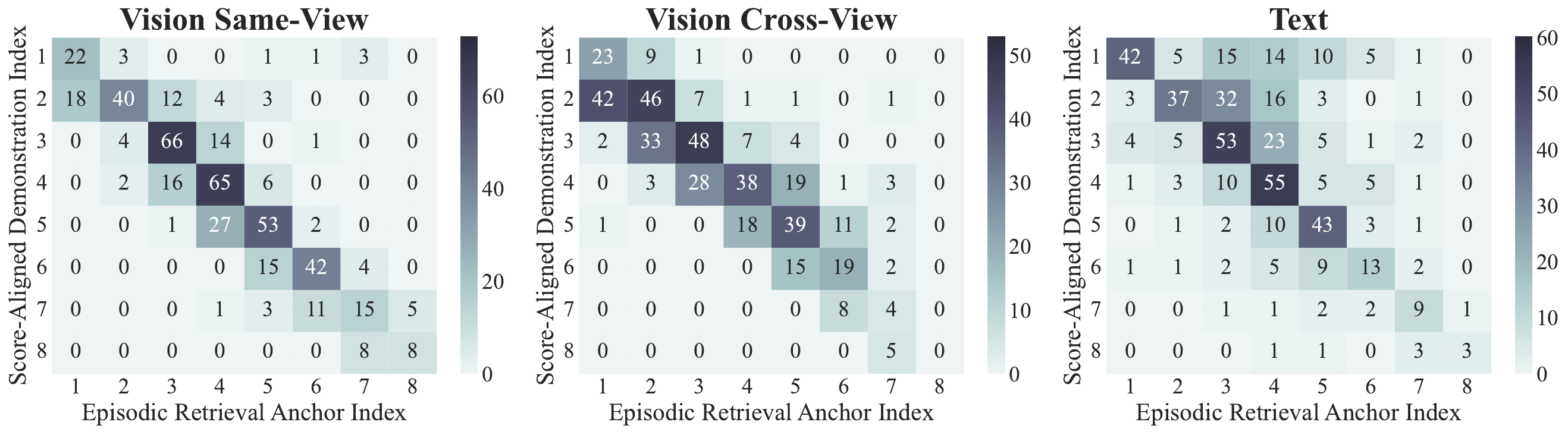}
    \caption{\textbf{Diagnostic analysis of coupled progress reasoning.} Heatmaps show the relationship between the episodic retrieval anchor index (x-axis) and the score-aligned demonstration index (y-axis) under Vision Same-View, Vision Cross-View, and Text settings. A strong diagonal indicates tight coupling between episodic retrieval and progress estimation. While coupling is strongest in the same-view setting and gradually weakens under cross-view and text conditions, the persistent diagonal structure across all settings demonstrates that progress estimation is consistently anchored to episodic retrieval rather than performed as direct regression.}
    \label{fig:app_heat}
\end{figure*}

\subsection{Chain-of-Thought Generation}

To construct high-quality Chain-of-Thought (CoT) training data for progress estimation, we employ a \textit{ground-truth guided generation} approach using Qwen2.5-VL-72B. Unlike conventional CoT distillation where models generate reasoning freely, our method constrains the generation by providing partial ground-truth answers—specifically the reference step index and final progress score—while requiring the model to synthesize coherent reasoning chains that justify these conclusions.

The generation process operates on two demonstration modalities:
(1) \textbf{Text-Based Demo CoT}: Given a task goal and step-by-step textual instructions (e.g., ``Step 1. reach for the power bank; Step 2. insert the battery...''), the model receives the current state image along with the ground-truth reference step (which text step most closely matches the current state) and progress score. The model generates reasoning in two phases: \texttt{<ref\_think>} explains why the given reference step is most relevant to the current image, and \texttt{<score\_think>} justifies how comparing the current state with the reference step yields the given progress score.
(2) \textbf{Visual-Based Demo CoT}: Given a sequence of demonstration frames with associated progress values (e.g., ``<image> 0\% <image> 25\% <image> 50\% <image> 75\% <image> 100\%''), the model receives the current state image and ground-truth annotations, then generates analogous reasoning explaining the visual comparison between the current state and demonstration frames.

This constrained generation strategy ensures that the synthesized CoT data exhibits consistent reasoning patterns aligned with correct answers, avoiding the noise introduced by freely-generated reasoning that may lead to incorrect conclusions.

\subsection{Human Bench Data}

To evaluate model generalization capabilities in real-world scenarios, we construct a human activity benchmark through manual data collection.

\noindent\textbf{Dataset Construction.}
Unlike existing robotic manipulation datasets collected from teleoperation systems, our benchmark captures \textit{human hand manipulation} activities in uncontrolled environments. Human annotators perform various manipulation tasks while recording video sequences from a top-down camera view. For each task, we collect: (1) a visual demonstration sequence showing the complete task execution, and (2) multiple test frames sampled from different execution instances of the same task. This setup creates natural domain shift between the demonstration and test frames, as different human performers exhibit variations in hand appearance, manipulation style, and environmental conditions.

\noindent\textbf{Task Categories.}
The human activity benchmark comprises 587 samples spanning 129 unique task goals across six manipulation categories:
(1) \textbf{Pushing}: moving objects toward target objects or positions (\emph{e.g.}, ``pushing a red cup to a rubic cube'');
(2) \textbf{Pick-and-Place}: placing objects into containers such as baskets (\emph{e.g.}, ``putting a jar into the blue basket'');
(3) \textbf{Placing on Surface}: positioning objects on flat surfaces like plates (\emph{e.g.}, ``placing an orange on a plate'');
(4) \textbf{Container Insertion}: inserting objects into enclosed containers (\emph{e.g.}, ``place the Rubik's cube inside the metal container'');
(5) \textbf{Stacking}: placing objects on top of other objects (\emph{e.g.}, ``putting a rubik's cube on the top of the box'');
(6) \textbf{Peg Manipulation}: Tower of Hanoi style block-on-peg tasks (\emph{e.g.}, ``place the blue block onto the middle red peg'').

\noindent\textbf{In-the-Wild Challenges.}
This benchmark introduces several challenges that push model capabilities beyond controlled laboratory settings:
(1) \textbf{Domain Gap}: models trained on robotic manipulation must generalize to human hand appearances and motion patterns;
(2) \textbf{Environmental Variation}: uncontrolled lighting, backgrounds, and object arrangements;
(3) \textbf{Execution Diversity}: different human performers exhibit distinct manipulation styles and trajectories;
(4) \textbf{Viewpoint Consistency}: top-down camera views differ from typical robotic camera setups.

\section{Experimental Settings}
\label{app:settings}

\subsection{Model Inference. }
For fair comparison, we adopt unified inference settings across all evaluated open-source vision-language models, including Qwen3-VL (4B, 8B, 32B), Qwen2.5-VL (3B, 7B, 32B, 72B), and InternVL3.5 (4B, 14B, 38B). Specifically, we set temperature to $0.6$, top-$p$ to $0.9$, and top-$k$ to $50$ for all models. The maximum number of generated tokens is set to 4,096 for inference. All models use bfloat16 precision and Flash Attention 2 for efficient inference.

\subsection{Text-Based Unanswerable Data Generation. } 
We use Qwen2.5-VL-72B \citep{bai2025qwen2} for text object replacement generation. The model is distributed across 4 NVIDIA H100 (80GB) GPUs using model parallelism. We set temperature to $0.7$, top-$p$ to $0.9$, top-$k$ to $50$, and maximum output tokens to $30{,}000$.

\subsection{Vision-Based Unanswerable Data Generation. } 
\noindent\textbf{Edit Prompt Generation. }We use Qwen2.5-VL-72B \citep{bai2025qwen2} for generating editing prompts. The model is loaded in bfloat16 precision with Flash Attention 2 \citep{dao2023flashattention} enabled. We set temperature to $0.7$ to encourage diverse editing strategies, top-$p$ to $0.9$, top-$k$ to $50$, and maximum output tokens to $1{,}024$. The image resolution is constrained between $1{,}003{,}520$ and $4{,}014{,}080$ pixels. Inference is conducted on 4 NVIDIA H100 (80GB) GPUs with a batch size of 8 per GPU. 

\noindent\textbf{Image Editing. }We use Qwen-Image-Edit (20B) \citep{wu2025qwenimagetechnicalreport} with 4 NVIDIA H100 (80GB) GPUs for image editing. Each GPU processes one image at a time (batch size of 1) due to the memory-intensive nature of diffusion models. The number of diffusion inference steps is set to $50$, and the classifier-free guidance scale is set to $4.0$. We use a single space character as the negative prompt and a fixed random seed of $42$ for reproducibility. All editing is performed in bfloat16 mixed precision.

\subsection{Chain-of-Thought Data Generation}

We use Qwen2.5-VL-72B \citep{bai2025qwen2} for CoT data generation with temperature 0.6, top-p 0.9, top-k 50, and maximum new tokens set to 4096 to accommodate extended reasoning chains. The model is distributed across 4 GPUs using model parallelism. For text-demo generation, we use batch size 40; for visual-demo generation with multiple input images per sample, we use batch size 8 to accommodate the increased memory requirements.

\subsection{Supervised Fine-Tuning. } We conduct Supervised Fine-Tuning (SFT) using the LLaMA-Factory framework~\citep{zheng2024llamafactory}. We adopt LoRA~\citep{hu2022lora} with rank 8 applied to all linear layers for parameter-efficient fine-tuning. The learning rate is set to $1 \times 10^{-4}$ with a cosine scheduler and $10\%$ warmup ratio. We use a per-device batch size of 2 with gradient accumulation steps of 8, resulting in an effective batch size of 64 on 4 NVIDIA H100 (80GB) GPUs. Models are trained for 2 epochs with BFloat16 mixed precision. The maximum sequence length is set to 30,000 tokens to accommodate multimodal inputs with multiple images.

\subsection{Reinforcement Learning. } We perform RL training using EasyR1 \citep{zheng2025easyr1}, a clean version of verl \citep{tao2025hybrid} with Group Relative Policy Optimization (GRPO)~\citep{shao2024deepseekmath} as the advantage estimator. The actor learning rate is $1 \times 10^{-6}$ with AdamW optimizer. We set the KL divergence coefficient to $0.01$ using the low-variance KL penalty. The global batch size is set to 64, with micro batch size of 1 per device for both policy update and experience collection. For each prompt, we generate $n=16$ rollout samples with temperature $0.6$ and top-$p$ of $0.9$. The maximum prompt length is 20,000 tokens and maximum response length is 8,192 tokens.  We use Fully Sharded Data Parallel (FSDP) \citep{zhao2023pytorch} and Ray \citep{moritz2018ray} for distributed training across 4 nodes with 4 NVIDIA H100 (80GB) GPUs per node (16 GPUs in total). The model is trained for 2 epochs with totally 20428 samples, requiring approximately 23 hours in total. We also using vLLM \citep{kwon2023efficient} serving as the generation engine with $60\%$ GPU memory utilization.

\section{Supplementary Results and Analysis}

\begin{table*}[t]
\centering
\caption{Human Bench: Comparison of in-the-wild model generalization performance on Visual and Textual Input evaluations.
\colorbox{first}{Best} \colorbox{second}{Second Best} \colorbox{third}{Third Best}. 
NSE$\downarrow$, PRC$\uparrow$, AFRR$\downarrow$. (think \textcolor{better}{better} or \textcolor{worse}{worse}).}
\label{tab:human_bench}
\setlength{\tabcolsep}{3pt}
\resizebox{\textwidth}{!}{
\begin{tabular}{lccccccccc}
\toprule
\multirow{2}{*}{\textbf{Model}} &
\multicolumn{3}{c}{\textbf{Vision-based Demo}} &
\multicolumn{3}{c}{\textbf{Text-based Demo}} &
\multicolumn{3}{c}{\textbf{Average}} \\
\cmidrule(lr){2-4} \cmidrule(lr){5-7} \cmidrule(lr){8-10}
& \textbf{NSE} & \textbf{PRC} & \textbf{AFRR}
& \textbf{NSE} & \textbf{PRC} & \textbf{AFRR}
& \textbf{NSE} & \textbf{PRC} & \textbf{AFRR} \\
\midrule
Qwen2.5VL-72B    & \cellcolor{third}23.5\% {\scriptsize\textcolor{better}{-1.8}} & \cellcolor{third}78.5\% {\scriptsize\textcolor{better}{+7.6}} & \cellcolor{third}0.2\% {\scriptsize\textcolor{better}{-0.1}}  & \cellcolor{first}23.8\% {\scriptsize\textcolor{worse}{+1.9}} & \cellcolor{first}75.3\% {\scriptsize\textcolor{worse}{-2.9}} & 1.9\% {\scriptsize\textcolor{worse}{+4.1}}  & \cellcolor{third}23.7\% {\scriptsize\textcolor{worse}{+0.1}} & \cellcolor{third}76.9\% {\scriptsize\textcolor{better}{+2.4}} & 1.1\% {\scriptsize\textcolor{worse}{+2.0}} \\
Qwen2.5VL-32B    & 31.7\% {\scriptsize\textcolor{better}{-4.3}} & 50.0\% {\scriptsize\textcolor{better}{+22.7}} & \cellcolor{first}0.0\%  & 26.7\% {\scriptsize\textcolor{worse}{+0.6}} & 70.5\% {\scriptsize\textcolor{worse}{-1.4}} & \cellcolor{second}0.2\% {\scriptsize\textcolor{worse}{+0.2}}  & 29.2\% {\scriptsize\textcolor{better}{-1.8}} & 60.2\% {\scriptsize\textcolor{better}{+10.6}} & 0.1\% {\scriptsize\textcolor{worse}{+0.1}} \\
Qwen2.5VL-7B     & 29.0\% {\scriptsize\textcolor{worse}{+11.0}} & 52.2\% {\scriptsize\textcolor{worse}{-11.0}} & 0.6\% {\scriptsize\textcolor{worse}{+21.7}}  & 29.3\% {\scriptsize\textcolor{worse}{+6.1}} & 58.7\% {\scriptsize\textcolor{worse}{-3.7}} & 11.4\% {\scriptsize\textcolor{better}{-4.4}}  & 29.2\% {\scriptsize\textcolor{worse}{+8.5}} & 55.4\% {\scriptsize\textcolor{worse}{-7.3}} & 6.0\% {\scriptsize\textcolor{worse}{+8.6}} \\
Qwen2.5VL-3B     & 27.8\% {\scriptsize\textcolor{worse}{+7.7}} & 48.0\% {\scriptsize\textcolor{worse}{-12.1}} & \cellcolor{first}0.0\% {\scriptsize\textcolor{worse}{+3.3}}  & 52.7\% {\scriptsize\textcolor{better}{-7.3}} & 17.2\% {\scriptsize\textcolor{better}{+11.2}} & 23.8\% {\scriptsize\textcolor{better}{-12.8}}  & 40.3\% {\scriptsize\textcolor{worse}{+0.2}} & 32.6\% {\scriptsize\textcolor{worse}{-0.4}} & 11.9\% {\scriptsize\textcolor{better}{-4.7}} \\
Qwen3VL-32B      & \cellcolor{second}19.7\% {\scriptsize\textcolor{worse}{+1.6}} & \cellcolor{first}91.7\% {\scriptsize\textcolor{worse}{-0.7}} & \cellcolor{first}0.0\% {\scriptsize\textcolor{worse}{+0.1}}  & \cellcolor{second}23.4\% {\scriptsize\textcolor{worse}{+0.7}} & \cellcolor{second}77.9\% {\scriptsize\textcolor{better}{+0.5}} & \cellcolor{third}0.1\% {\scriptsize\textcolor{worse}{+0.2}}  & \cellcolor{second}21.5\% {\scriptsize\textcolor{worse}{+1.1}} & \cellcolor{first}84.8\% {\scriptsize\textcolor{worse}{-0.1}} & \cellcolor{first}0.04\% {\scriptsize\textcolor{worse}{+0.1}} \\
Qwen3VL-8B       & \cellcolor{second}19.6\% {\scriptsize\textcolor{worse}{+4.3}} & \cellcolor{second}80.9\% {\scriptsize\textcolor{better}{+3.3}} & \cellcolor{first}0.0\% {\scriptsize\textcolor{worse}{+0.7}}  & 25.0\% {\scriptsize\textcolor{worse}{+0.3}} & 69.3\% {\scriptsize\textcolor{better}{+3.8}} & \cellcolor{third}0.1\% {\scriptsize\textcolor{worse}{+4.8}}  & 22.3\% {\scriptsize\textcolor{worse}{+2.3}} & 75.1\% {\scriptsize\textcolor{better}{+3.5}} & \cellcolor{first}0.04\% {\scriptsize\textcolor{worse}{+2.8}} \\
Qwen3VL-4B       & 22.4\% {\scriptsize\textcolor{worse}{+1.6}} & 80.4\% {\scriptsize\textcolor{better}{+1.3}} & \cellcolor{first}0.0\% {\scriptsize\textcolor{worse}{+0.9}}  & 25.7\% {\scriptsize\textcolor{better}{-0.9}} & 68.5\% {\scriptsize\textcolor{better}{+0.9}} & \cellcolor{first}0.0\% {\scriptsize\textcolor{worse}{+1.1}}  & 24.0\% {\scriptsize\textcolor{worse}{+0.3}} & 74.4\% {\scriptsize\textcolor{better}{+1.1}} & \cellcolor{first}0.0\% {\scriptsize\textcolor{worse}{+1.0}} \\
Qwen3VL-2B       & 48.4\% {\scriptsize\textcolor{worse}{+2.8}} & 33.6\% {\scriptsize\textcolor{better}{+11.6}} & \cellcolor{third}0.1\% {\scriptsize\textcolor{worse}{+13.2}}  & 67.4\% {\scriptsize\textcolor{better}{-17.6}} & 5.6\% {\scriptsize\textcolor{better}{+32.2}} & \cellcolor{first}0.0\% {\scriptsize\textcolor{worse}{+10.5}}  & 57.9\% {\scriptsize\textcolor{better}{-7.4}} & 19.6\% {\scriptsize\textcolor{better}{+21.9}} & \cellcolor{first}0.04\% {\scriptsize\textcolor{worse}{+11.9}} \\
Intern3.5-VL-38B & 40.8\% {\scriptsize\textcolor{worse}{+17.6}} & 44.5\% {\scriptsize\textcolor{worse}{-8.9}} & 14.2\% {\scriptsize\textcolor{better}{-10.3}} & \cellcolor{third}25.9\% {\scriptsize\textcolor{better}{-0.1}} & \cellcolor{third}59.7\% {\scriptsize\textcolor{better}{+2.0}} & 7.7\% {\scriptsize\textcolor{better}{-3.3}} & 33.3\% {\scriptsize\textcolor{worse}{+8.7}} & 52.1\% {\scriptsize\textcolor{worse}{-3.4}} & 10.9\% {\scriptsize\textcolor{better}{-6.8}} \\
Intern3.5-VL-14B & 34.4\% {\scriptsize\textcolor{worse}{+24.8}} & 52.2\% {\scriptsize\textcolor{worse}{-24.9}} & 8.0\% {\scriptsize\textcolor{better}{-7.8}} & 30.0\% {\scriptsize\textcolor{better}{-1.4}} & 53.8\% {\scriptsize\textcolor{better}{+3.2}} & 0.6\% {\scriptsize\textcolor{worse}{+1.2}}  & 32.2\% {\scriptsize\textcolor{worse}{+11.7}} & 53.0\% {\scriptsize\textcolor{worse}{-10.9}} & 4.3\% {\scriptsize\textcolor{better}{-3.3}} \\
Intern3.5-VL-4B  & 34.7\% {\scriptsize\textcolor{worse}{+15.2}} & 50.3\% {\scriptsize\textcolor{worse}{-20.1}} & \cellcolor{first}0.0\% {\scriptsize\textcolor{worse}{+0.1}}  & 33.6\% {\scriptsize\textcolor{better}{-4.5}} & 35.3\% {\scriptsize\textcolor{better}{+16.7}} & 0.9\% {\scriptsize\textcolor{worse}{+0.6}} & 34.2\% {\scriptsize\textcolor{worse}{+5.4}} & 42.8\% {\scriptsize\textcolor{worse}{-1.7}} & 0.5\% {\scriptsize\textcolor{worse}{+0.4}} \\
ProgressLM-SFT-3B & \cellcolor{second}19.7\% & 76.4\% & 3.2\%  & 32.4\% & 49.6\% & 5.8\%  & 26.0\% & 63.0\% & 4.5\% \\
ProgressLM-RL-3B  & \cellcolor{first}15.5\% & \cellcolor{first}88.9\% & 0.9\%  & 30.9\% & 46.0\% & 11.3\%  & \cellcolor{first}23.2\% & \cellcolor{second}67.5\% & 6.1\% \\
\bottomrule
\end{tabular}
}
\end{table*}

\begin{figure*}[ht]
    \centering
    \includegraphics[width=\linewidth]{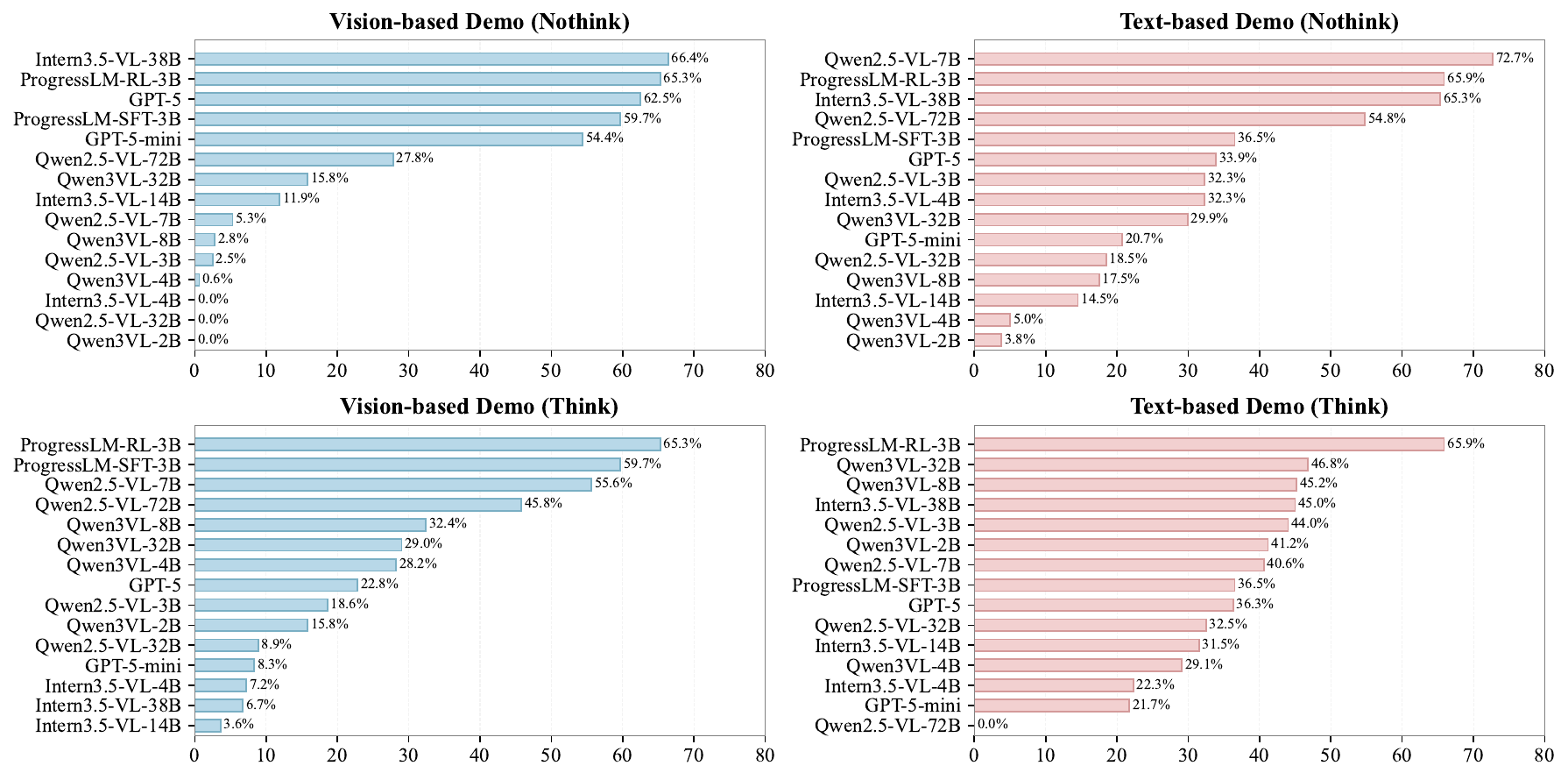}
    \caption{\textbf{Unanswerable Detection Accuracy (UDA) across models with and without training-free thinking.} This figure compares unanswerable detection accuracy under Vision-based and Text-based demonstrations, contrasting standard inference (NoThink) with training-free explicit reasoning (Think). Across both modalities, enabling training-free thinking consistently improves UDA for most models, with particularly pronounced gains in text-based settings where semantic mismatch is harder to identify. The results highlight that explicit reasoning at inference time enhances models’ ability to recognize ill-defined progress and correctly abstain, complementing the benefits brought by our training-based coupled reasoning approach.
    }
    \label{fig:nega_barchart_all}
\end{figure*}

\subsection{Vision-Based Demo Case Studies.}
\label{app:vision_case}

We provide qualitative case studies to further illustrate how vision-based demonstrations support coupled progress reasoning under both same-view and cross-view settings. These examples complement the quantitative results reported in the main paper and offer mechanistic insights into how episodic retrieval and progress estimation interact in practice.

\paragraph{Same-view reasoning with fine-grained state alignment.}
In the same-view case, the demonstration frames and the state to estimate share a consistent viewpoint, enabling direct visual alignment. As shown in the plate-stacking example, the model retrieves a demonstration step corresponding to a near-completion stage, where the plates are almost fully stacked. Progress estimation is then performed by comparing subtle state differences, such as the remaining motion of the robot arm and the degree of object contact. This behavior aligns with the strong diagonal structure observed in the Vision Same-View diagnostic heatmaps and the low NSE reported in the corresponding benchmark results, indicating that progress estimation is tightly anchored to the retrieved episodic reference.

\paragraph{Robust reasoning under cross-view variations.}
The cross-view case presents a more challenging setting, where the observation is captured from a viewpoint different from the demonstration sequence. In the block relocation example, despite significant viewpoint changes, the model successfully retrieves a semantically aligned anchor step representing a late-stage placement of the block. Progress is estimated by reasoning over task-relevant state changes, such as object position and the robot’s disengagement from manipulation, rather than relying on pixel-level similarity. The resulting prediction closely matches the ground truth, reflecting the softer but still structured coupling between retrieval and estimation observed in the Vision Cross-View setting. This qualitative behavior is consistent with the broader diagonal patterns and slightly increased NSE seen in cross-view evaluations.

\paragraph{Connection to quantitative trends.}
These cases help explain the performance gap between vision-based and text-based demonstrations observed across benchmarks. Vision-based demos provide dense and continuous state information, allowing the model to ground episodic retrieval in physical states and perform local mental simulation for progress estimation. This grounding leads to higher PRC and lower AFRR compared to text-based inputs, as confirmed by the quantitative results. Even under viewpoint changes, vision-based demonstrations preserve sufficient semantic structure to support reliable progress reasoning.

\paragraph{Key insight.}
Together, these visualizations reveal that effective progress estimation relies on retrieving a semantically aligned visual anchor and reasoning locally around that reference. Same-view settings enable near-deterministic coupling between retrieval and estimation, while cross-view settings introduce uncertainty that weakens but does not break this coupling. These qualitative findings provide concrete evidence that vision-based demonstrations play a critical role in enabling robust, generalizable progress reasoning, especially under domain shifts and partial observability.

\begin{figure*}[ht]
    \centering
    \includegraphics[width=\linewidth]{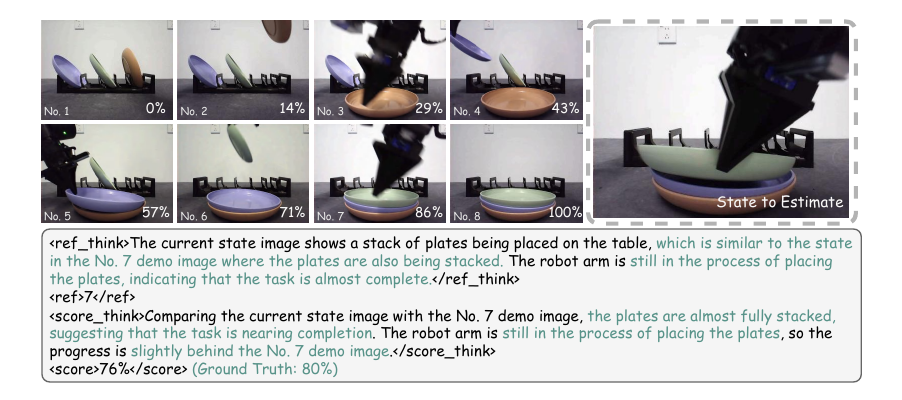}
    \caption{\textbf{Vision-Based Case Visualization (Same-View)}. This example illustrates how the model performs progress estimation by coupling episodic retrieval with mental simulation. Given a current observation (right), the model retrieves the most semantically aligned demonstration step (No. 7) from the visual demo sequence (left), where the plates are nearly stacked. Based on this retrieved anchor, the model estimates the relative progress by comparing fine-grained state differences, yielding a progress prediction of 76\% against a ground-truth of 80\%. The intermediate reasoning explicitly shows how reference selection and score estimation are jointly grounded in the demonstration sequence.
    }
    \label{fig:visual_same}
\end{figure*}

\begin{figure*}[ht]
    \centering
    \includegraphics[width=\linewidth]{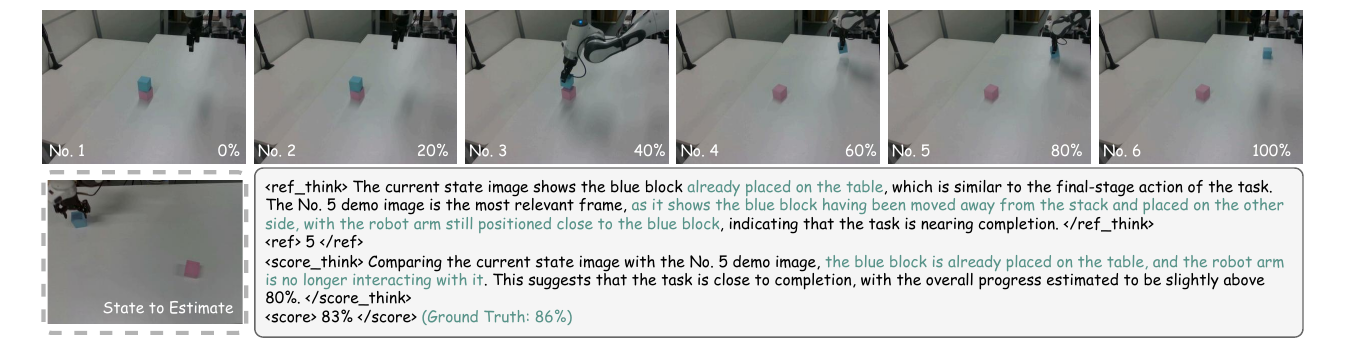}
    \caption{\textbf{Vision-Based Case Visualization (Cross-View)}. This example illustrates progress estimation with cross-view visual demonstrations, where the observation is captured from a viewpoint different from the demonstration sequence. Given the current state image (bottom left), the model retrieves the most semantically aligned demonstration step (No. 5), corresponding to a near-completion stage where the blue block has been placed on the table. By comparing fine-grained state differences between the retrieved anchor and the current observation, the model estimates the task progress as 83\%, closely matching the ground truth of 86\%. This case highlights the model’s ability to perform robust episodic retrieval and progress estimation despite viewpoint changes.
    }
    \label{fig:visual_cross}
\end{figure*}

\begin{figure*}[ht]
    \centering
    \includegraphics[width=\linewidth]{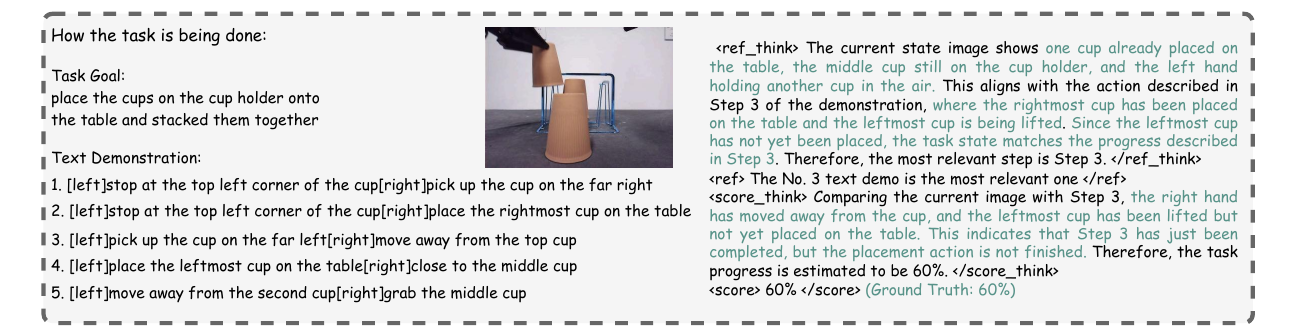}
    \caption{\textbf{Text-Based Case Visualization}. This example illustrates progress estimation with text-based demonstrations. Given the current visual observation, the model retrieves the most semantically aligned textual step (Step 3) from the demonstration by matching object states and action semantics described in language. Progress is then estimated by reasoning over the partially completed placement action, resulting in a prediction of 60\%, which exactly matches the ground truth. This case highlights how episodic retrieval over textual steps can effectively anchor progress estimation, even when demonstrations are provided purely in language.
    }
    \label{fig:text_case}
\end{figure*}

\begin{figure*}[ht]
    \centering
    \includegraphics[width=\linewidth]{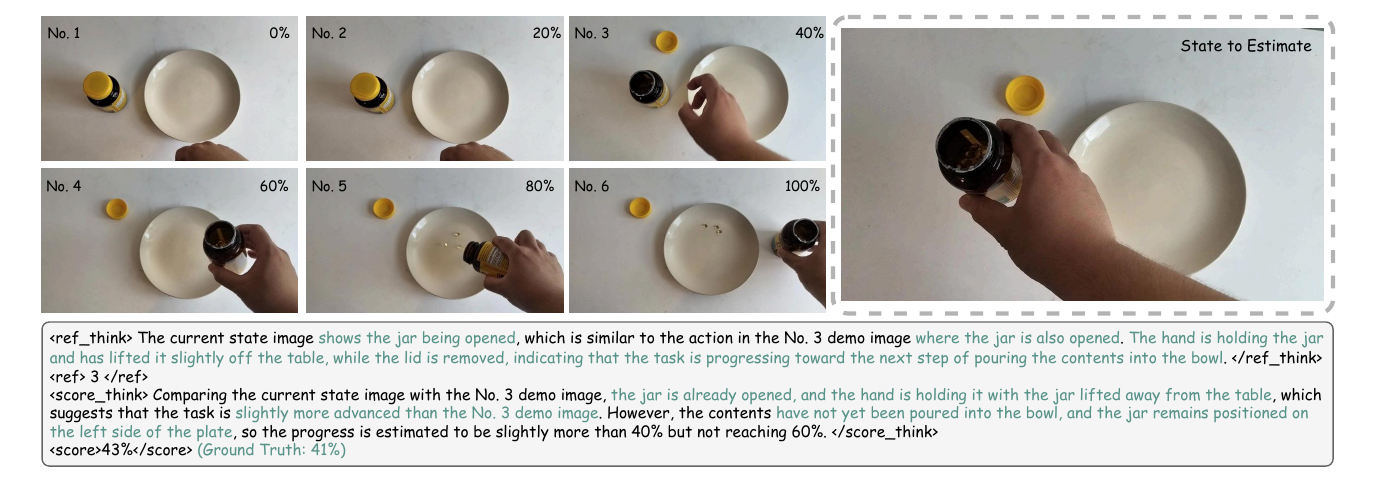}
    \caption{\textbf{In-the-wild Generalization on Human Activities}. This example demonstrates the model’s ability to generalize coupled progress reasoning beyond robotic manipulation to human-performed activities. Given a sequence of demonstration frames depicting the step-by-step process of opening a jar and pouring its contents, the model retrieves the most semantically aligned demonstration step (No. 3) for the current observation and estimates the task progress by comparing subtle state differences. The predicted progress (43\%) closely matches the ground truth (41\%), illustrating that episodic retrieval and progress estimation remain effective in unconstrained, real-world human activity scenarios.
    }
    \label{fig:visual_human}
\end{figure*}

\begin{figure*}[ht]
    \centering
    \includegraphics[width=\linewidth]{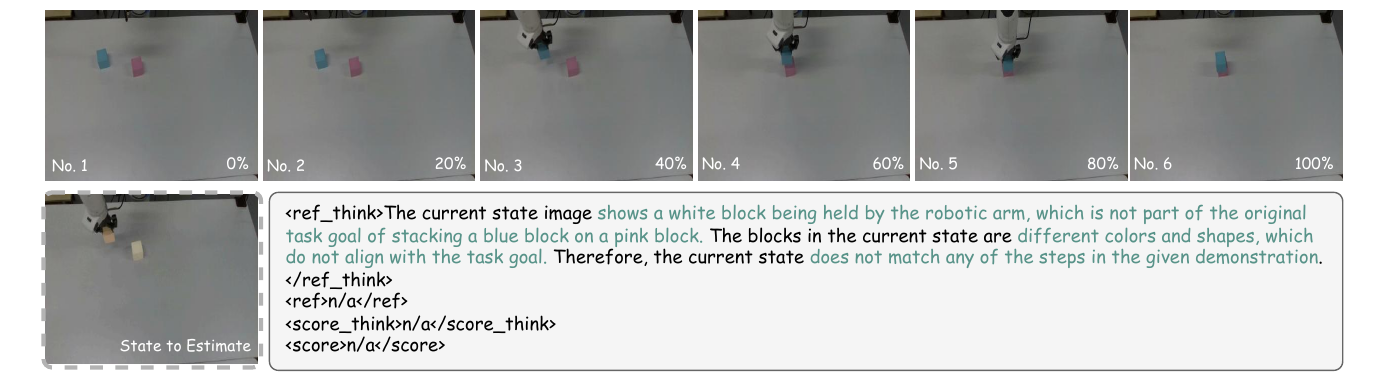}
    \caption{\textbf{Visual Unanswerable Case Visualization}. This example illustrates a visual unanswerable scenario where the current observation is semantically inconsistent with the given demonstration. While the demonstration depicts a task of stacking a blue block on a pink block, the observed state shows the robot holding an unrelated white block that does not appear in any demonstration step. As no valid episodic anchor can be retrieved and progress estimation is ill-defined, the model correctly abstains by predicting N/A. This case highlights the model’s ability to detect semantic mismatch and avoid spurious progress estimation.
    }
    \label{fig:visual_nega_case}
\end{figure*}

\begin{figure*}[ht]
    \centering
    \includegraphics[width=\linewidth]{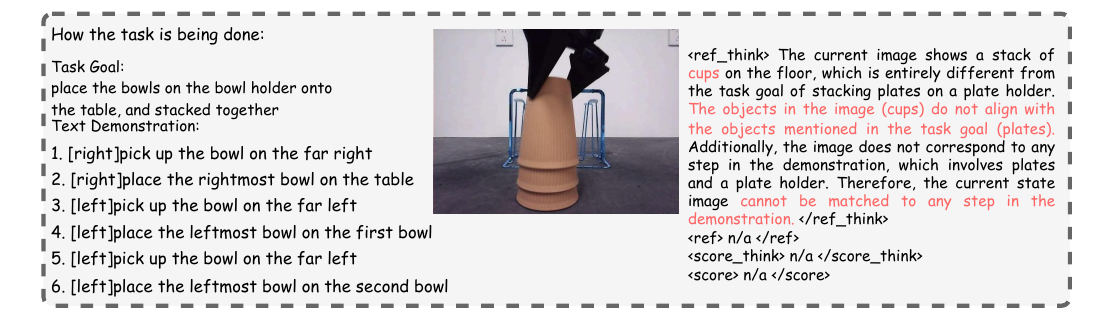}
    \caption{\textbf{Text Unanswerable Case Visualization}. This example illustrates a text unanswerable scenario where the current visual observation is semantically incompatible with the textual demonstration. While the task goal and instructions describe stacking bowls on a bowl holder, the observed state contains a stack of cups on the floor, involving different object categories and spatial configurations. As the observation cannot be aligned with any textual step in the demonstration, episodic retrieval fails and progress estimation becomes ill-defined, leading the model to correctly output N/A. This case highlights the model’s ability to detect cross-modal semantic mismatch and abstain from spurious progress predictions.
    }
    \label{fig:text_nega_case}
\end{figure*}

\subsection{Text-Based Demo Case Studies.}
\label{app:text_case}

We present qualitative case studies to analyze progress estimation with text-based demonstrations. Compared to vision-based demos, text-only instructions provide abstract and discrete descriptions of task execution, requiring the model to ground linguistic steps to visual observations before estimating progress.

\paragraph{Episodic retrieval from textual steps.}
In the cup-stacking example shown in Figure \ref{fig:text_case}, the model first performs episodic retrieval over the textual demonstration and identifies Step~3 as the most semantically aligned reference. This step describes a transitional state where one cup has been placed on the table while another is being lifted. The retrieved anchor reflects a correct alignment between the linguistic action description and the observed physical state, despite the absence of explicit visual cues in the demonstration itself.

\paragraph{Progress estimation under abstract supervision.}
Conditioned on the retrieved textual anchor, the model estimates progress by reasoning about which sub-actions have been completed and which remain unfinished. In this case, the leftmost cup has been lifted but not yet placed, indicating that the task has just completed Step~3 but has not transitioned to Step~4. Accordingly, the model predicts a progress of 60\%, exactly matching the ground truth. This behavior demonstrates that progress estimation is anchored to the relative position within the textual sequence rather than inferred directly from global visual appearance.

\paragraph{Relation to quantitative results.}
These observations help explain the quantitative trends reported in the main paper. Text-based demonstrations consistently yield higher NSE and AFRR compared to vision-based inputs, reflecting increased ambiguity in episodic retrieval when multiple visual states correspond to the same textual instruction. Nevertheless, the successful alignment in this example shows that when the textual anchor is correctly identified, the model can still perform accurate progress estimation through structured reasoning.

\paragraph{Key insight.}
These case studies reveal that text-based progress estimation remains feasible but inherently more sensitive to retrieval errors. Effective reasoning depends on accurately grounding abstract textual steps to observed physical states, after which progress estimation can be performed through local comparison within the retrieved episode. This further supports our view that progress estimation, even under purely textual supervision, benefits from an explicit coupling between episodic retrieval and mental simulation.

\subsection{Analysis of Coupled Progress Two-Stage Reasoning}
\label{app: coupled reasoning}

To further diagnose whether progress estimation is performed as a coupled reasoning process, we analyze the interaction between \textbf{episodic retrieval} and \textbf{progress score prediction} at an intermediate level. Specifically, we examine whether the demonstration step selected as the episodic anchor is aligned with the step that is semantically consistent with the predicted progress score.

For each test instance, we record: (1) the \textbf{episodic retrieval anchor index}, corresponding to the demonstration step selected by the model as reference; and (2) the \textbf{score-aligned demonstration index}, defined as the step whose ground-truth progress interval best matches the model’s predicted score. We aggregate these pairs into a 2D histogram, shown as heatmaps (See Figure \ref{fig:app_heat}) under two settings: Vision-Based Demonstration (Same-View and Cross-View), and Text-Based Demonstration.

\noindent\textbf{Vision Same-View.} Under the vision same-view setting, the heatmap exhibits a clear and sharp diagonal structure, indicating strong alignment between the retrieved anchor and the score-consistent demonstration step. This suggests that when visual states are well aligned, the model reliably retrieves the correct episodic reference and performs progress estimation within its local context. The tight concentration along the diagonal provides strong evidence that progress prediction is not performed as an isolated regression, but is explicitly anchored to episodic retrieval.

\noindent\textbf{Vision Cross-View.} In the cross-view setting, the diagonal structure remains evident but becomes noticeably wider, with increased mass in neighboring indices. This reflects higher uncertainty in episodic retrieval caused by viewpoint changes, where multiple demonstration steps may be visually or semantically plausible anchors. Importantly, the distribution still concentrates around the diagonal, indicating that progress estimation remains conditioned on episodic retrieval, albeit in a softer and less deterministic manner.

\noindent\textbf{Text.} For text-based demonstrations, the heatmap shows the weakest alignment, with broader dispersion across indices. This behavior is expected, as textual steps often correspond to abstract or overlapping physical states, making episodic retrieval inherently more ambiguous. Nevertheless, the persistence of a diagonal trend indicates that even in the absence of visual grounding, the model continues to estimate progress relative to a retrieved textual anchor rather than collapsing into direct score regression.

In summary, across all settings, these results consistently demonstrate that progress estimation emerges as a \textbf{coupled} reasoning process, where \textbf{\textit{episodic retrieval serves as a prerequisite for mental simulation and score estimation}}. The gradual degradation from vision same-view to cross-view and text highlights how modality-induced uncertainty weakens but does not remove this coupling, providing mechanistic evidence that supports our model design and training strategy.

\subsection{In the Wild Generalization Analysis}
\label{subsec:in_the_wild}

We analyze in-the-wild generalization using \textsc{Human-Bench}, which evaluates progress estimation on human-performed activities under unconstrained settings. Compared to robotic demonstrations, these scenarios introduce substantial domain shifts in embodiment, motion dynamics, object appearance, and execution variability, making reliable progress estimation and abstention behavior significantly more challenging.

\paragraph{Overall trends.}
As shown in Table~\ref{tab:human_bench}, most vision-language models exhibit clear degradation in \textbf{NSE} and \textbf{PRC} under in-the-wild conditions, accompanied by elevated \textbf{AFRR}. This indicates that human activities amplify both progress miscalibration (higher NSE) and incorrect abstention behavior (higher AFRR), especially when progress must be inferred from subtle hand–object interactions rather than rigid robot motions.

\paragraph{Impact of model scale and visual grounding.}
Larger models with stronger visual grounding, such as Qwen3VL-32B, consistently achieve higher \textbf{PRC} and near-zero \textbf{AFRR}, suggesting improved recognition of valid progress states. However, their \textbf{NSE} remains relatively high, indicating that increased model capacity alone is insufficient to ensure fine-grained progress calibration in human activities. Smaller variants (e.g., Qwen3VL-2B and Qwen2.5VL-7B) suffer from both elevated NSE and reduced PRC, reflecting compounded errors in episodic alignment and progress estimation.

\paragraph{Vision versus text demonstrations.}
Across nearly all models, vision-based demonstrations outperform text-based ones in terms of both \textbf{NSE} and \textbf{PRC}. Text-based inputs consistently yield higher \textbf{AFRR}, revealing a tendency to incorrectly abstain when step boundaries are ambiguous. This behavior aligns with the abstract nature of textual steps in human activities, where multiple physical states may correspond to a single instruction, weakening episodic retrieval and downstream progress estimation.

\paragraph{Effectiveness of coupled progress reasoning.}
Despite its smaller scale, \textsc{ProgressLM-RL-3B} achieves the lowest average \textbf{NSE} while maintaining competitive \textbf{PRC} and controlled \textbf{AFRR}. Compared to the SFT-only variant, reinforcement learning consistently reduces NSE, indicating improved calibration of continuous progress estimates. \textbf{\textit{These gains suggest that explicitly coupling episodic retrieval with progress estimation is particularly beneficial under domain shift, where robust anchor selection becomes critical}}.

\paragraph{Qualitative alignment with in-the-wild cases.}
The quantitative trends are consistent with the qualitative example in Figure~14. In the jar-opening task, the model retrieves a semantically aligned demonstration step corresponding to the jar being opened and estimates progress by comparing fine-grained state differences, resulting in a prediction (43\%) closely matching the ground truth (41\%). This example illustrates how accurate episodic anchoring enables stable progress estimation even in the presence of human-specific variability.

\paragraph{Key insight.}
These results suggest that in-the-wild generalization depends less on raw model capacity and more on whether progress estimation is performed as a structured, coupled reasoning process. Models that fail to anchor progress estimation to semantically aligned episodic references tend to exhibit higher \textbf{NSE} and \textbf{AFRR}, while \textsc{ProgressLM} demonstrates that explicitly modeling this coupling leads to more reliable progress reasoning beyond curated robotic environments.

\subsection{Unanswerable Case Recognition}
\label{app:unanswerable_recognition}

We analyze the ability of models to recognize unanswerable cases, where progress estimation is ill-defined due to semantic mismatch between demonstrations and observations. This capability is critical for safe and reliable progress reasoning, as erroneous score prediction in such cases leads to spurious confidence and degraded downstream performance.

\paragraph{Training-free thinking improves unanswerable recognition.}
Figure~13 compares unanswerable detection accuracy (UDA) across models under standard inference (\emph{NoThink}) and training-free explicit reasoning (\emph{Think}). A consistent pattern emerges across both vision-based and text-based demonstrations: enabling explicit reasoning at inference time substantially improves UDA for most models. The improvement is especially pronounced in text-based settings, where semantic mismatch is more abstract and harder to detect from surface cues alone.

\paragraph{Vision versus text demonstrations.}
Under vision-based demonstrations, several large models already achieve moderate UDA in the NoThink setting, suggesting that visual inconsistencies such as object category or spatial violations can often be detected via perceptual cues. However, training-free thinking further improves performance by encouraging explicit comparison between the current state and retrieved demonstration steps, leading to more reliable abstention decisions. In contrast, text-based demonstrations exhibit much lower NoThink performance, indicating that without explicit reasoning, models tend to hallucinate progress scores even when no textual step aligns with the observation. The gains from Think in this setting highlight the importance of structured semantic comparison for detecting cross-modal inconsistency.

\paragraph{Model-dependent effects and limitations of scale.}
While larger models generally benefit more from training-free thinking, the results reveal that model scale alone does not guarantee robust unanswerable recognition. Several large models still exhibit limited UDA under NoThink inference, particularly for text-based inputs. This suggests that recognizing unanswerable cases requires not only capacity but also an explicit reasoning process that verifies the existence of a valid episodic anchor before attempting progress estimation.

\paragraph{Interaction with training-based coupled reasoning.}
Notably, models trained with our coupled progress reasoning framework, such as \textsc{ProgressLM}, demonstrate strong UDA even without training-free thinking, and further benefit when Think is enabled. This indicates a complementary relationship between training-based supervision and inference-time reasoning: training encourages the model to internalize the prerequisite that progress estimation depends on successful episodic retrieval, while training-free thinking makes this dependency explicit during inference. Together, they lead to more robust detection of ill-defined progress scenarios.

\paragraph{Key insight.}
These findings suggest that unanswerable case recognition is fundamentally a reasoning problem rather than a purely perceptual one. Reliable detection requires verifying whether a semantically aligned episodic reference exists before estimating progress. Training-free thinking provides a lightweight mechanism to expose this verification process at inference time, while training-based coupled reasoning reinforces it structurally. The combination of both yields the most reliable unanswerable recognition across modalities and model scales.

\begin{figure*}[t]
    \centering
    \includegraphics[width=1\linewidth]{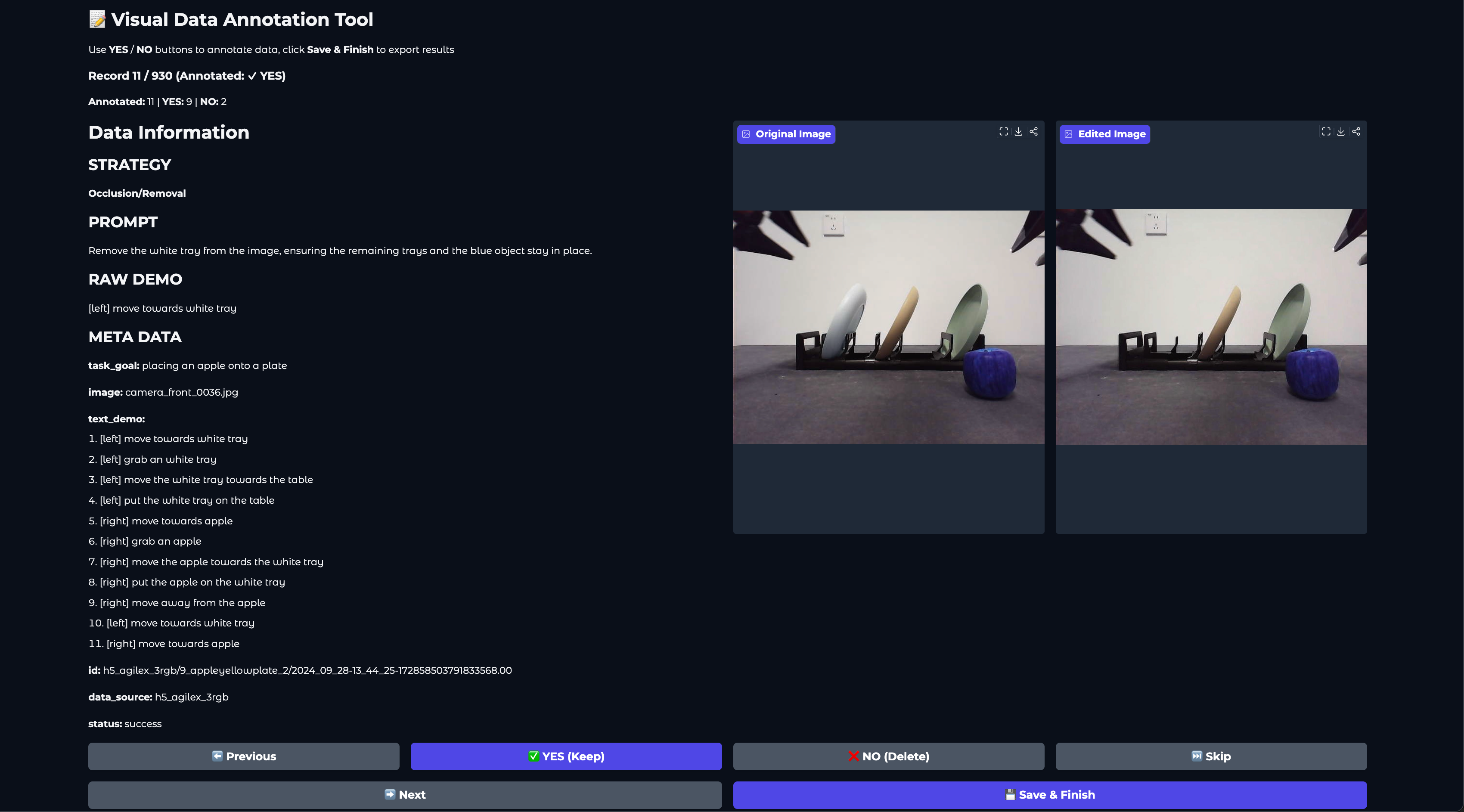}
    \caption{\textbf{Gradio-based Human Filtering Platform for Visual Unanswerable Data Generation.} We employ a Gradio-based annotation interface to manually verify the quality of edited images used for visual unanswerable construction. Annotators are presented with the original and edited images alongside the task goal, step-level demonstrations, editing strategy, and prompt. Each edited sample is retained only if it simultaneously violates the intended manipulation step and preserves visual realism, ensuring high-quality and reliable visual unanswerable data.}
    \label{fig:gradio}
\end{figure*}

\section{Prompts}

\subsection{Vision-based Demo}

\begin{table*}[htbp]
\centering
\begin{tcolorbox}[
    colframe=black,
    colback=gray!10!white,
    coltitle=black,
    boxrule=0.5mm,
    enhanced
]

You are a progress estimator that evaluates the progress of the current state during an ongoing task based on a visual demonstration. The demonstration consists of a sequence of vision-based states and their corresponding progress value (ranging from 0\% to 100\%), showing how the task evolves from start to completion.
\bigskip

\textbf{Here is the demonstration:}\par
[Insert the ordered set of demonstration frames, representing sequential progress from earliest stage to latest stage]
\bigskip

\textbf{Here is the current state that you need to estimate:}\par
[Insert the single image ``stage\_to\_estimate'']
\bigskip

\textbf{Your task:}
\begin{enumerate}
    \item Check the current state image carefully.
    \item Analyze the overall task goal and visual demonstration to understand how the task progresses from start to completion.
    \item Identify the reference states from the visual demonstration that are most related to the current state image.
    \item Compare the current state image with the chosen reference state, determining whether the image is behind or after the reference state.
    \item Estimate the progress numerically as a floating-point value between 0\% and 100\%.
    \item If you really cannot match the current state image to any of the states from demonstration, you need to explain the reason within `<ref\_think></ref\_think>` and output "n/a" within `<ref></ref>`, `<score\_think></score\_think>`, and `<score></score>`.
\end{enumerate}
\bigskip

\textbf{Your response must strictly follow this format:}

\verb|<ref_think>| Reason for choosing the most related state from the demonstration as the reference or explanation of why the current state image does not match the task goal or any steps from demonstration \verb|</ref_think>|\par

\verb|<ref>| which state from the visual demonstration is most related to the current state (output only the number of the state) or "n/a" \verb|</ref>|\par

\verb|<score_think>| Reason for comparing the current state image with the reference state or "n/a" \verb|</score_think>|\par

\verb|<score>| Your final estimated progress score or "n/a" \verb|</score>|

\end{tcolorbox}
\caption{Prompt for Visual Demo Inference}
\label{prompt:visual_infer}
\end{table*}

\subsection{Text-based Demo}

\begin{table*}[htbp]
\centering
\begin{tcolorbox}[colframe=black, colback=gray!10!white, coltitle=black, boxrule=0.5mm, enhanced]

You are a progress estimator that evaluates the progress of the current state during an ongoing task based on a textual demonstration. The demonstration consists of a sequence of text-based steps and their corresponding progress value (ranging from 0\% to 100\%), showing how the task evolves from start to completion.
\bigskip

\textbf{Here is the demonstration:}\par
[Insert the full ordered text\_demo containing all steps and their associated progress values]
\bigskip

\textbf{Here is the current state that you need to estimate:}\par
[Insert the single image named ``stage\_to\_estimate'']
\bigskip

\textbf{Your task:}
\begin{enumerate}
    \item Read the task goal to understand the task objective and the entity being operated on.
    \item Analyze the textual demonstration to understand how the task progresses from start to completion.
    \item Examine the current state image carefully. If the target is incorrect (different from the object metioned in task goal) or you really cannot match the current image to any step in the demonstration, you must explain the reason within \verb|<ref_think></ref_think>| and output ``n/a'' within \verb|<ref></ref>|, \verb|<score_think></score_think>|, and \verb|<score></score>|.
    \item If a match is possible, examine all steps in the textual demonstration, where each step represents an independent action. Identify the single step whose action is most closely related to the current state image. Then compare the current image with that reference step to determine whether it corresponds to an earlier or later stage, and finally estimate the overall progress as a floating-point value between 0\% and 100\%.
\end{enumerate}
\bigskip

\textbf{Your response must strictly follow this format:}

\verb|<ref_think>| Explain the reason for selecting the most relevant step from the demonstration. If the task target is incorrect, or the current state image cannot be matched to any demonstration step, explain why here. \verb|</ref_think>|\par

\verb|<ref>| If a valid matching step exists, output only the step number. If the task target is incorrect or no step matches the current image, output only ``n/a''. Please ensure that this is the same as the ref value you reasoned before. \verb|</ref>|\par

\verb|<score_think>| If a valid matching step exists, explain how you compare the current image with that step to judge progress. If the task target is incorrect or no step matches the current image, output only ``n/a''. \verb|</score_think>|\par

\verb|<score>| If a valid matching step exists, output the estimated progress score (0\%--100\%). If the task target is incorrect or no step matches the current image, output only ``n/a''. \verb|</score>|

\end{tcolorbox}
\caption{Prompt for Text Demo Inference}
\label{prompt:text_infer}
\end{table*}

\subsection{Vision-based Chain-of-Thought Prompt}

\begin{table*}[htbp]
\centering
\begin{tcolorbox}[colframe=black, colback=gray!10!white, coltitle=black, boxrule=0.5mm, enhanced]

\small

You are an expert AI analyst specializing in generating step-by-step reasoning for visual task-progress evaluations. Your objective is not to estimate from scratch. Instead, your task is to construct a perfect, human-like chain of thought that logically explains and justifies a known, ground-truth progress score. Your entire response must read as if you are deducing the conclusion independently from visual analysis alone.

You are a progress estimator specializing in evaluating the progress of an ongoing task based on visual evidence. The demonstration consists of a sequence of video frames (images) showing how the task evolves from 0\% (start) to 100\% (completion). Your goal is to produce a human-like reasoning chain that logically supports the given progress score.
\medskip

\textbf{Here is the demonstration:}\par
[Insert the ordered set of demo images representing progress stages, from early to late]
\medskip

\textbf{Here is the current state that you need to estimate:}\par
[Insert the single image named ``stage\_to\_estimate'']
\medskip

\textbf{Critical Rule}\par
The correct final progress score will be provided to you. However, you must never reveal or imply that you already know the answer. Your reasoning must appear as a fully original, independent visual analysis derived from the images.
\medskip

\textbf{Ground-Truth Progress Result}\par
Closest Reference Frame: \verb|{closest_idx_str}| \\
Final Progress Score to Justify: \verb|{progress_score_str}| 
\medskip

\textbf{Abnormal Situation Handling:}\par
If you detect any of the following abnormal situations:
\begin{itemize}
    \item The current state does not match the task goal or any visual demo images
    \item The operation appears to have failed or resulted in an error state
\end{itemize}

You must output ``n/a'' for both \verb|<ref>| and \verb|<score>|. In your reasoning sections, clearly explain why the situation is abnormal and why no valid progress estimation can be made.
\medskip

\textbf{Your task:}
\begin{enumerate}
    \item Analyze the demonstration images to understand how the task visually progresses from start to completion.
    \item Identify the frame (or frames) from the demonstration that are visually most similar to the current state image.
    \item Compare the current state to that reference frame and determine whether it shows more or less progress.
    \item Finally, provide a numeric progress estimation between 0\% and 100\%, or both \verb|<ref>| and \verb|<score>| be ``n/a'' while encountering abnormal situation.
\end{enumerate}
\medskip

\textbf{Your response must strictly follow this format:}
\medskip

\verb|<ref_think>| Your reasoning for choosing the closest demonstration frame as the reference, OR explanation of why the situation is abnormal and no reference can be identified \verb|</ref_think>|\par
\medskip
\verb|<ref>| The progress score of your chosen reference frame, OR ``n/a'' if abnormal situation detected \verb|</ref>|\par
\medskip
\verb|<score_think>| Your reasoning for comparing the current state image with the reference frame, OR explanation of why no valid progress score can be assigned \verb|</score_think>|\par
\medskip
\verb|<score>| Your final estimated progress score, OR ``n/a'' if abnormal situation detected \verb|</score>|

\end{tcolorbox}
\vspace{-1em}
\caption{Chain-of-Thought Construction for Vision-Based Demo}
\label{prompt:visual_cot}
\end{table*}

\subsection{Text-based Chain-of-Thought Prompt}

\begin{table*}[htbp]
\centering
\begin{tcolorbox}[colframe=black, colback=gray!10!white, coltitle=black, boxrule=0.5mm, enhanced]

\small

You are an expert AI analyst specializing in visual task-progress evaluations. Your objective is not to estimate from scratch. Instead, your task is to construct a perfect, human-like chain of thought that logically explains and justifies a known, ground-truth progress score. Your entire response must read as if you are deducing the conclusion independently from visual analysis alone.
\medskip

This is the system prompt for normal inference. You are a progress estimator that evaluates the progress of an ongoing task based on a textual demonstration of its step-by-step progression. The demonstration consists of a sequence of text instructions (text\_demo), each describing one step of the process. Each step explicitly states the corresponding progress value (ranging from 0\% to 100\%), showing how the task evolves from start to completion.
\medskip

\textbf{Here is the demonstration:}\par
[Insert the full ordered text\_demo containing all steps and their associated progress values]
\medskip

\textbf{Here is the current state that you need to estimate:}\par
[Insert the single image named ``stage\_to\_estimate'']
\medskip

\textbf{Critical Rule}\par
The correct final progress score will be provided to you. However, you must \textbf{never} reveal or imply that you already know the answer. Your reasoning must appear as a fully original, independent visual analysis derived from the images.
\medskip

\textbf{Ground-Truth Progress Result}\par
Closest Reference Frame: The No. \verb|{closest_idx}| text demo is the most relevant one \\
Final Progress Score to Justify: \verb|{final_progress_score}|
\medskip

\textbf{Abnormal Situation Handling:}\par
If you detect any of the following abnormal situations:
\begin{itemize}
    \item The current state does not match the task goal or any demo steps
    \item The operation appears to have failed or resulted in an error state
\end{itemize}
You must output ``n/a'' for both \verb|<ref>| and \verb|<score>|. In your reasoning sections, clearly explain why the situation is abnormal and why no valid progress estimation can be made.
\medskip

\textbf{Your task:}
\begin{enumerate}
    \item Analyze the text\_demo to understand how the task visually and conceptually progresses from start to completion.
    \item Identify the step from the text\_demo that are most visually and semantically similar to the current state image.
    \item Compare the current state image with the chosen reference step to determine whether it represents an earlier or later stage.
    \item Estimate the progress numerically as a floating-point value between 0\% and 100\%, or both \verb|<ref>| and \verb|<score>| be ``n/a'' while encontering abnormal situation.
\end{enumerate}
\medskip

\textbf{Your response must strictly follow this format:}
\medskip

\verb|<ref_think>| Your reasoning for choosing the most similar text\_demo step as the reference, OR explanation of why the situation is abnormal and no reference can be identified \verb|</ref_think>|\par
\medskip
\verb|<ref>| which text demo is most semantically similar to the current state (output only the number), OR ``n/a'' if abnormal situation detected \verb|</ref>|\par
\medskip
\verb|<score_think>| Your reasoning for comparing the current state image with the reference step, OR explanation of why no valid progress score can be assigned \verb|</score_think>|\par
\medskip
\verb|<score>| Your final estimated progress score, OR ``n/a'' if abnormal situation detected \verb|</score>|

\end{tcolorbox}
\vspace{-1em}
\caption{Chain-of-Thought Construction for Text-Based Demo}
\label{prompt:text_cot}
\end{table*}

\subsection{Unanswerable Vision-based Sample Generation}

\begin{table*}[htbp]
\centering
\begin{tcolorbox}[colframe=black, colback=gray!10!white, coltitle=black, boxrule=0.5mm, enhanced]

\small

You are tasked with constructing adversarial image edits that intentionally cause failure in an instruction-following, multi-step visual manipulation task while preserving realism and coherence. Your goal is to make the provided image no longer align with its corresponding step instruction.
\medskip

\textbf{Input Information:}\par
Task Goal: \verb|{task_goal}|\par
Step-by-step Instructions: \verb|{text_demo}|\par
Current Image: [The provided image]\par
Corresponding Instruction: Step \verb|{step_number}| -- {specific\_instruction}
\medskip

\textbf{Your Task:}\par
You are given an image that corresponds to a specific step in a multi-step robotic manipulation task. Your goal is to edit this image to make it no longer align with the corresponding instruction, causing the task to fail.
\medskip

\textbf{Editing Guidelines:}\par
Modify key objects or elements in the image using one of the following strategies:
\begin{enumerate}
    \item Color Change: Alter the color of critical objects (e.g., change a red apple to green)
    \item Object Replacement: Replace the target object with a different object (e.g., replace an egg with an orange)
    \item Occlusion/Removal: Hide or remove key objects from the scene
\end{enumerate}
\medskip

\textbf{Requirements:}\par
\begin{enumerate}
    \item The edited image should clearly violate the corresponding instruction.
    \item Maintain visual realism and coherence—the edited image must look natural and believable.
    \item Ensure the edit would cause the overall task goal to fail.
    \item The modification should be semantically meaningful (not just noise or blur).
\end{enumerate}
\medskip

\textbf{Output Format:}\par
\medskip

\verb|<strategy_think>| Analyze the current instruction and image content. Think step by step about which editing strategy would most effectively violate this instruction while maintaining realism. Consider the key objects involved and how modifying them would break the instruction. \verb|</strategy_think>|\par
\medskip

\verb|<strategy>| State the single strategy you selected from the editing guidelines (e.g., "Object Replacement" or "Color Change") \verb|</strategy>|\par
\medskip

\verb|<prompt_think>| Think step by step about how to formulate a clear and effective image editing prompt. Consider: What specific change to make? Which objects to target? What details are needed for realism? \verb|</prompt_think>|\par
\medskip

\verb|<prompt>| Write a concise image editing prompt (maximum 20 words) that clearly instructs the editing model what to change in the image. \verb|</prompt>|

\end{tcolorbox}
\caption{Adversarial Image Editing Prompt Generation for Unanswerable Visual Samples}
\label{prompt:visual_nega}
\end{table*}

\subsection{Unanswerable Text-Based Sample Generation}

\begin{table*}[htbp]
\centering
\begin{tcolorbox}[colframe=black, colback=gray!10!white, coltitle=black, boxrule=0.5mm, enhanced]

Task: Task: Modify the Task Goal and Step-by-step Instructions to make the Current Image does not match the Task Goal or any Step-by-step Instructions.
\bigskip

\textbf{Input Information:}\par
- Task Goal: {task\_goal}\par
- Step-by-step Instructions: {text\_demo}\par
- Current Image: [The provided image]\par
\bigskip

\textbf{Editing Guidelines:}\par
1. Keep the original sentence format and structure - ONLY replace the object name.\par
2. For each step in Step-by-step Instructions, preserve ALL markers like [right], [left], [towards], etc. in their EXACT original positions.\par
\bigskip

\textbf{Output Format:}\par
\verb|<edited_goal>| "put your edited task goal here" \verb|</edited_goal>|\par

\verb|<edited_demo>|\par
"text\_demo": ["your edited step 1", "your edited step 2", "your edited step 3", ..., "your edited step n"]\par
\verb|</edited_demo>|\par

\end{tcolorbox}
\caption{Object Replacement for Unanswerable Text Sample Generation}
\label{prompt:text_nega}
\end{table*}

\end{document}